\documentclass{article} % For LaTeX2e
\usepackage[accepted]{icml2018}
\usepackage{times}
\usepackage[colorlinks=true,citecolor=blue]{hyperref}
\usepackage{url}
\usepackage{bm}
\usepackage{microtype}
\usepackage{booktabs}
\usepackage{comment}
\usepackage{amsmath}
\usepackage{amssymb}
\usepackage{graphicx}
\usepackage{subcaption}
\usepackage{balance}

\hypersetup{
  colorlinks = true,
  citecolor = blue,
  pdftitle={Convolutional Normalizing Flows},
  pdfauthor={Guoqing Zheng, Yiming Yang, Jaime Carbonell}
}

\newcommand{\vect}[1]{\bm{#1}}

\icmltitlerunning{Convolutional Normalizing Flows}

\begin{document}

\twocolumn[
\icmltitle{Convolutional Normalizing Flows}



 
\begin{icmlauthorlist}
  \icmlauthor{Guoqing Zheng}{cmu}
  \icmlauthor{Yiming Yang}{cmu}
  \icmlauthor{Jaime Carbonell}{cmu}
\end{icmlauthorlist}

\icmlaffiliation{cmu}{School of Computer Science, Carnegie Mellon University, Pittsburgh PA, USA}
\icmlcorrespondingauthor{Guoqing Zheng}{gzheng@cs.cmu.edu}

\icmlkeywords{Machine Learning, ICML}

\vskip 0.3in
]



\printAffiliationsAndNotice{}  % leave blank if no need to mention equal contribution

%\iclrfinalcopy % Uncomment for camera-ready version, but NOT for submission.

\graphicspath{{plots/}}

\begin{abstract}
  
Bayesian posterior inference is prevalent in various machine learning
problems. Variational inference provides one way to approximate the
posterior distribution, however its expressive power is limited and so
is the accuracy of resulting approximation. Recently, there has a
trend of using neural networks to approximate the variational
posterior distribution due to the flexibility of neural network
architecture. One way to construct flexible variational distribution
is to warp a simple density into a complex by normalizing flows, where
the resulting density can be analytically evaluated. However, there is
a trade-off between the flexibility of normalizing flow and
computation cost for efficient transformation. In this paper, we
propose a simple yet effective architecture of normalizing flows,
\textit{ConvFlow}, based on convolution over the dimensions of random
input vector. Experiments on synthetic and real world posterior
inference problems demonstrate the effectiveness and efficiency of the
proposed method.
\end{abstract}

\section{Introduction}

Posterior inference is the key to Bayesian modeling, where we are
interested to see how our belief over the variables of interest change
after observing a set of data points. Predictions can also benefit
from Bayesian modeling as every prediction will be equipped with
confidence intervals representing how sure the prediction is. Compared
to the maximum a posterior estimator of the model parameters, which
is a point estimator, the posterior distribution provide richer
information about the model parameter hence enabling more justified
prediction.

Among the various inference algorithms for posterior estimation,
variational inference (VI) and Monte Carlo Markov chain (MCMC) are the
most two widely used ones. It is well known that MCMC suffers from
slow mixing time though asymptotically the samples from the chain will
be distributed from the true posterior. VI, on the other hand,
facilitates faster inference, since it is optimizing an explicit
objective function and convergence can be measured and controlled, and
it's been widely used in many Bayesian models, such as Latent
Dirichlet Allocation \citep{DBLP:journals/jmlr/BleiNJ03}, etc. However,
one drawback of VI is that it makes strong assumption about the shape
of the posterior such as the posterior can be decomposed into multiple
independent factors. Though faster convergence can be achieved by
parameter learning, the approximating accuracy is largely limited.

The above drawbacks stimulates the interest for richer function
families to approximate posteriors while maintaining acceptable
learning speed. Specifically, neural network is one among such models
which has large modeling capacity and endows efficient
learning. \citep{DBLP:conf/icml/RezendeM15} proposed normalization
flow, where the neural network is set up to learn an invertible
transformation from one known distribution, which is easy to sample
from, to the true posterior. Model learning is achieved by minimizing
the KL divergence between the empirical distribution of the generated
samples and the true posterior. After properly trained, the model
will generate samples which are close to the true posterior, so that
Bayesian predictions are made possible. Other methods based on
modeling random variable transformation, but based on different
formulations are also explored, including NICE~\citep{dinh2014nice},
the Inverse Autoregressive Flow \citep{DBLP:conf/nips/KingmaSJCCSW16},
and Real NVP~\citep{dinh2016density}.

One key component for normalizing flow to work is to compute the
determinant of the Jacobian of the transformation, and in order to
maintain fast Jacobian computation, either very simple function is
used as the transformation, such as the planar flow
in~\citep{DBLP:conf/icml/RezendeM15}, or complex tweaking of the
transformation layer is required. Alternatively, in this paper we
propose a simple and yet effective architecture of normalizing flows,
based on convolution on the random input vector. Due to the nature of
convolution, bi-jective mapping between the input and output vectors
can be easily established; meanwhile, efficient computation of the
determinant of the convolution Jacobian is achieved linearly. We
further propose to incorporate dilated convolution~\citep{yu2015multi,
  oord2016wavenet} to model long range interactions among the input
dimensions. The resulting convolutional normalizing flow, which we
term as \textit{Convolutional Flow (ConvFlow)}, is simple and yet
effective in warping simple densities to match complex ones.

The remainder of this paper is organized as follows: We briefly review
the principles for normalizing flows in Section \ref{sec:prelim}, and
then present our proposed normalizing flow architecture based on
convolution in Section \ref{sec:convflow}. Empirical evaluations and
analysis on both synthetic and real world data sets are carried out in
Section \ref{sec:exp}, and we conclude this paper in Section
\ref{sec:conclusion}.

\section{Preliminaries}
\label{sec:prelim}

\subsection{Transformation of random variables}

Given a random variable $\vect z\in\mathbb{R}^d$ with density $p(\vect
z)$, consider a smooth and invertible function
$f:\mathbb{R}^d\rightarrow\mathbb{R}^d$ operated on $\vect z$. Let
$\vect z'=f(\vect z)$ be the resulting random variable,  the
density of $\vect z'$ can be evaluated as
\begin{align}
  p(\vect z')=p(z)\left|\det \frac{\partial f^{-1}}{\partial \vect z'}\right|=p(z)\left|\det \frac{\partial f}{\partial \vect z}\right|^{-1}
\end{align}
thus
\begin{align}
  \log p(\vect z')=\log p(z)-\log \left|\det \frac{\partial f}{\partial \vect z}\right|
  \end{align}

\subsection{Normalizing flows}
 
Normalizing flows considers successively transforming $\vect z_0$ with a
series of transformations $\{f_1, f_2, ..., f_K\}$ to construct
arbitrarily complex densities for $\vect z_K=f_K\circ f_{K-1}\circ...\circ f_1(\vect z_0)$ as
\begin{align}
  \log p(\vect z_K)=\log p(\vect z_0)-\sum_{k=1}^K\log\left|\det \frac{\partial f_k}{\partial \vect z_{k-1}}\right|
\end{align}

Hence the complexity lies in computing the determinant of the Jacobian
matrix. Without further assumption about $f$, the general complexity
for that is $\mathcal{O}(d^3)$ where $d$ is the dimension of $\vect
z$. In order to accelerate this, \citep{DBLP:conf/icml/RezendeM15} proposed the following family
of transformations that they termed as \textit{planar flow}:
\begin{align}
  f(\vect z)=\vect z+\vect uh(\vect w^\top \vect z+ b)
\end{align}
where $\vect w\in\mathbb{R}^d, \vect u\in\mathbb{R}^d, b\in\mathbb{R}$
are parameters and $h(\cdot)$ is a univariate non-linear function with
derivative $h'(\cdot)$. For this family of transformations, the
determinant of the Jacobian matrix can be computed as
\begin{align}
  \det\frac{\partial f}{\partial \vect z}=\det(\vect I+\vect u\psi(\vect z)^\top)=1+\vect u^\top \psi(\vect z)
\end{align}
where $\psi(\vect z)=h'(\vect w^\top \vect z+b)\vect w$. The
computation cost of the determinant is hence reduced from
$\mathcal{O}(d^3)$ to $\mathcal{O}(d)$.

Applying $f$ to $\vect z$ can be viewed as feeding the input variable
$\vect z$ to a neural network with only one single hidden unit
followed by a linear output layer which has the same dimension with
the input layer. Obviously, because of the bottleneck caused by the
single hidden unit, the capacity of the family of transformed density
is hence limited. 

\section{A new transformation unit}
\label{sec:convflow}

In this section, we first propose a general extension to the above
mentioned planar normalizing flow, and then propose a restricted
version of that, which actually turns out to be convolution over the
dimensions of the input random vector.
%transformation unit which admits larger distribution coverage as well as maintaining linear computation efficiency.

\subsection{Normalizing flow with $d$ hidden units}

Instead of having a single hidden unit as suggested in planar flow,
consider $d$ hidden units in the process. We denote the weights
associated with the edges from the input layer to the output layer as
$\vect W\in\mathbb{R}^{d\times d}$ and the vector to adjust the magnitude of each dimension of
the hidden layer activation as $\vect u$, and the transformation is
defined as
\begin{align}
  f(\vect z)=\vect u \odot h(\vect W\vect z+\vect b)
\end{align}
where $\odot$ denotes the point-wise multiplication. The Jacobian matrix of
this transformation is
\begin{align}
  \frac{\partial f}{\partial \vect z}&=\mbox{diag}(\vect u\odot h'(\vect W\vect z+b))\vect W\\
  \det\frac{\partial f}{\partial \vect z}&=\det[\mbox{diag}(\vect u \odot h'(\vect W\vect z+b))]\det(\vect W)
  \end{align}

As $\det(\mbox{diag}(\vect u\odot h'(\vect W\vect z+\vect b)))$ is
linear, the complexity of computing the above transformation lies in
computing $\det (\vect W)$. Essentially the planar flow is restricting
$\vect W$ to be a vector of length $d$ instead of matrices, however we
can relax that assumption while still maintaining linear complexity of
the determinant computation based on a very simple fact that the
determinant of a triangle matrix is also just the product of the
elements on the diagonal.

\subsection{Convolutional Flow}

Since normalizing flow with a fully connected layer may not be
bijective and generally requires $\mathcal{O}(d^3)$ computations for
the determinant of the Jacobian even it is, we propose to use 1-d
convolution to transform random vectors.
\begin{figure}[!htb]
  \centering
  \begin{subfigure}{0.385\linewidth}
    \includegraphics[width=\textwidth]{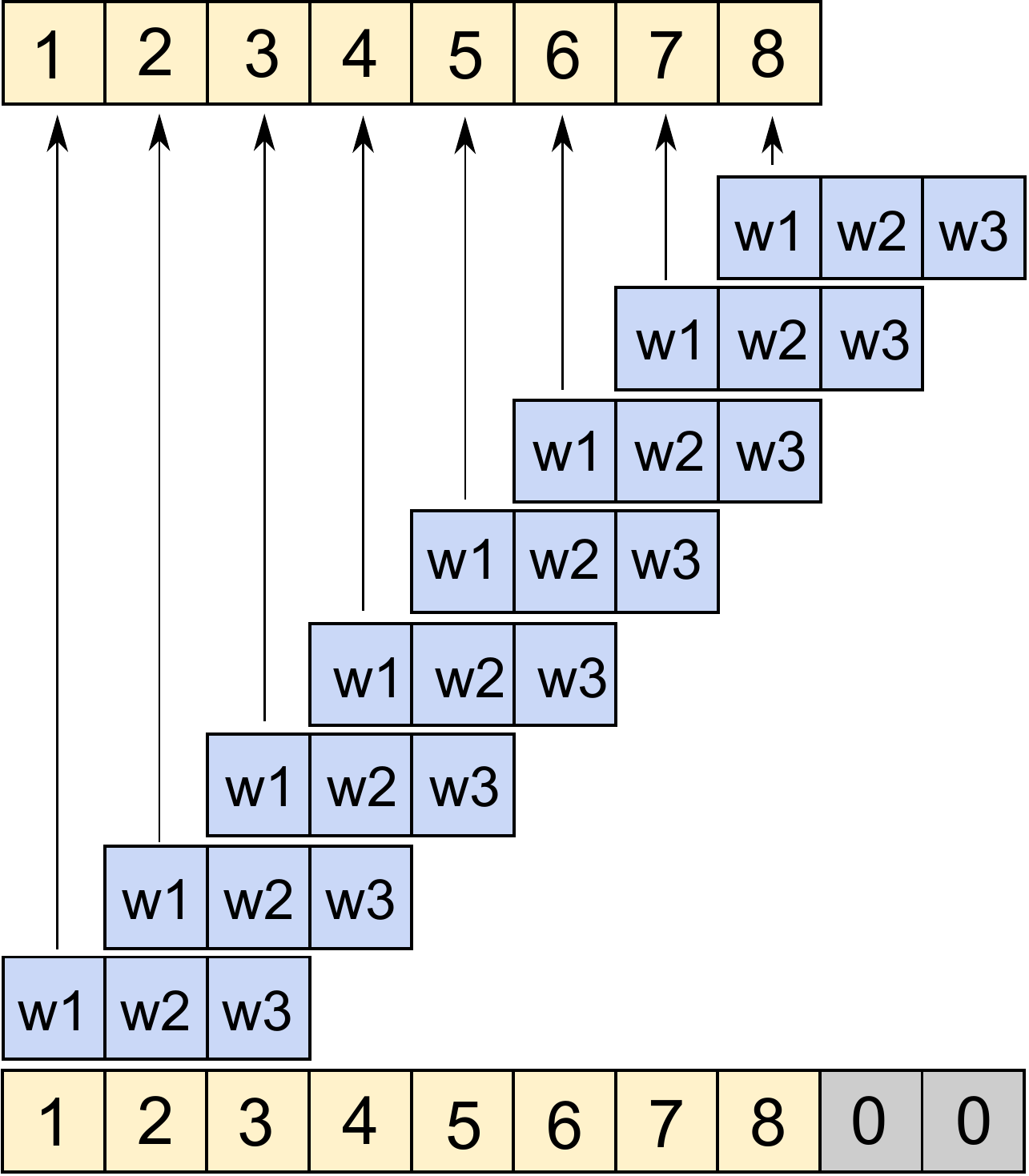}
    \caption{}
  \end{subfigure}
  \hfill
  \begin{subfigure}{0.595\linewidth}
    \includegraphics[width=\textwidth]{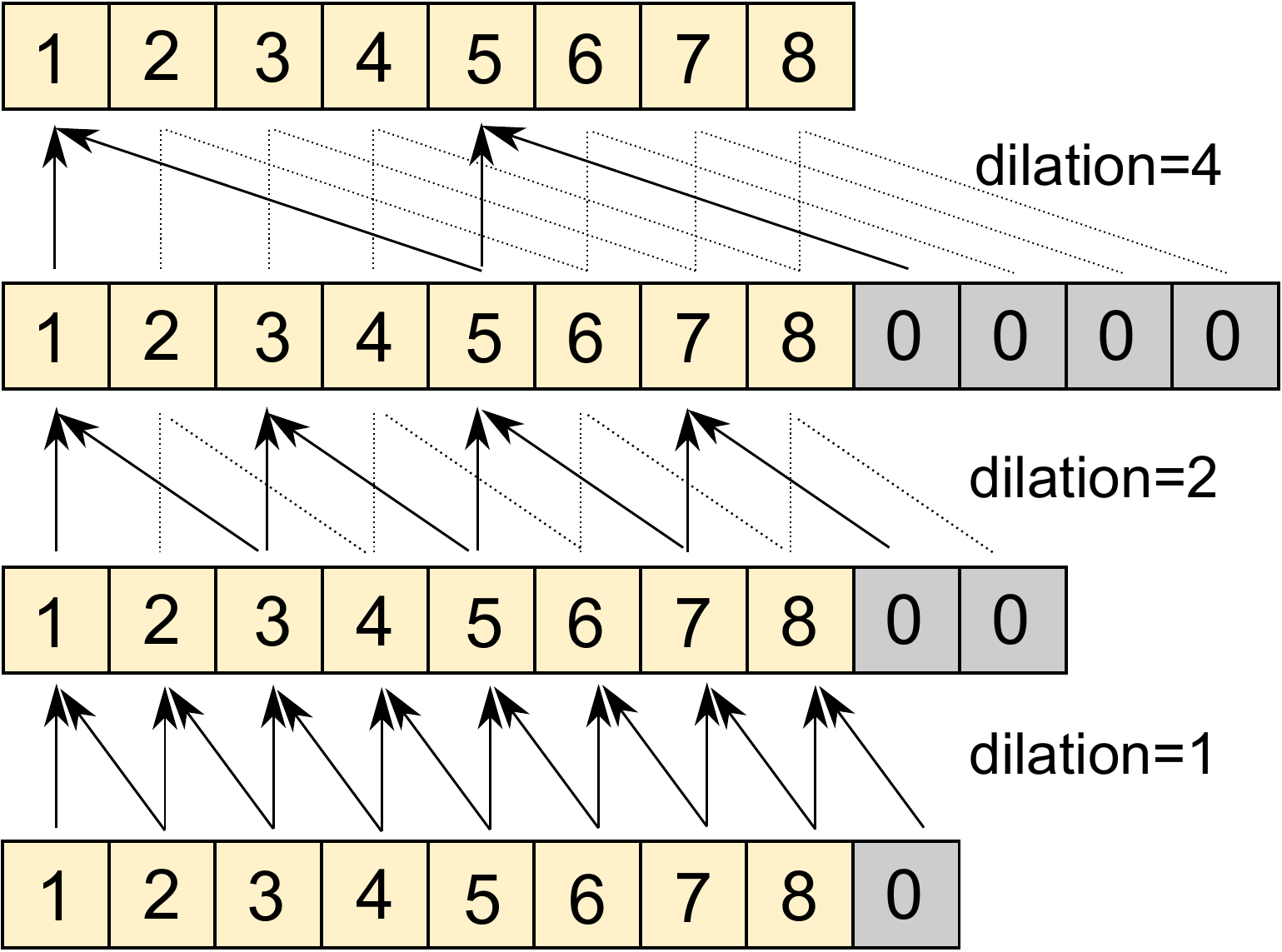}
    \caption{}
  \end{subfigure}
  \caption{(a) Illustration of 1-D convolution, where the dimensions
    of the input/output variable are both 8 (the input vector is
    padded with 0), the width of the convolution filter is 3 and
    dilation is 1; (b) A block of ConvFlow layers stacked with
    different dilations.}
  \label{fig:1dconv}
\end{figure}

Figure \ref{fig:1dconv}(a) illustrates how 1-d convolution is
performed over an input vector and outputs another vector. We propose
to perform a 1-d convolution on an input random vector $\vect z$,
followed by a non-linearity and necessary post operation after
activation to generate an output vector. Specifically,
%.l, specifically
\begin{align}
  f(\vect z)=\vect z + \vect u\odot h(\mbox{conv}(\vect z, \vect w))
\end{align}
where $\vect w\in\mathbb{R}^k$ is the parameter of the 1-d convolution
filter ($k$ is the convolution kernel width), $\mbox{conv}(\vect
z,\vect w)$ is the 1d convolution operation as shown in Figure
\ref{fig:1dconv}(a), $h(\cdot)$ is a monotonic non-linear activation
function\footnote{Examples of valid $h(x)$ include all conventional
  activations, including sigmoid, tanh, softplus, rectifier (ReLU),
  leaky rectifier (Leaky ReLU) and exponential linear unit (ELU).},
$\odot$ denotes point-wise multiplication, and $\vect
u\in\mathbb{R}^d$ is a vector adjusting the magnitude of each
dimension of the activation from $h(\cdot)$. We term this normalizing
flow as \textit{Convolutional Flow (ConvFlow)}.

ConvFlow enjoys the following properties
\begin{itemize}
\item Bi-jectivity can be easily achieved with standard and fast 1d
  convolution operator if proper padding and a monotonic activation
  function with bounded gradients are adopted (Minor care is needed to
  guarantee strict invertibility, see Appendix \ref{app:condition} for
  details);
\item Due to local connectivity, the Jacobian determinant of ConvFlow
  only takes $\mathcal{O}(d)$ computation independent from convolution
  kernel width $k$ since
\begin{align}
  \frac{\partial f}{\partial \vect z}=I+\mbox{diag}(w_1\vect u\odot
  h'(\mbox{conv}(\vect z, \vect w)))
  \label{eq:jacobian}
\end{align}
where $w_1$ denotes the first element of $\vect w$.  \\ For example
for the illustration in Figure \ref{fig:1dconv}(a), the Jacobian
matrix of the 1d convolution $\mbox{conv}(\vect z, \vect w)$ is
\begin{align}
  &\frac{\partial\,\mbox{conv}(\vect z, \vect w)}{\partial \vect z}\nonumber\\
  =&\begin{bmatrix}
    w_1 & w_2 & w_3 & &  & & &\\
    &  w_1 & w_2 & w_3 & &  & &\\
    &    &  w_1 & w_2 & w_3 & &  & \\
    &     &    &  w_1 & w_2 & w_3 & &   \\
    & &     &    &  w_1 & w_2 & w_3 &    \\
    & & &     &    &  w_1 & w_2 & w_3    \\
    && & &     &    &  w_1 & w_2     \\
    &&& & &     &    &  w_1      \\
  \end{bmatrix}
\end{align}
which is a triangular matrix whose determinant can be easily computed;
\item ConvFlow is much simpler than previously proposed variants of
  normalizing flows. The total number of parameters of one ConvFlow
  layer is only $d+k$ where generally $k<d$, particularly efficient
  for high dimensional cases. Notice that the number of parameters in
  the planar flow in~\citep{DBLP:conf/icml/RezendeM15} is $2d$ and one
  layer of Inverse Autoregressive Flow (IAF)
  \citep{DBLP:conf/nips/KingmaSJCCSW16} and Real
  NVP~\citep{dinh2016density} require even more parameters. In Section
  \ref{sec:compiaf}, we discuss the key differences of ConvFlow from
  IAF in detail.
\end{itemize}

A series of $K$ ConvFlows can be stacked to generate complex output
densities. Further, since convolutions are only visible to inputs from
adjacent dimensions, we propose to incorporate dilated
convolution~\citep{yu2015multi,oord2016wavenet} to the flow to
accommodate interactions among dimensions with long distance
apart. Figure \ref{fig:1dconv}(b) presents a block of 3 ConvFlows
stacked, with different dilations for each layer. Larger receptive
field is achieved without increasing the number of parameters. We term
this as a ConvBlock.

From the block of ConvFlow layers presented in Figure
\ref{fig:1dconv}(b), it is easy to verify that dimension $i\,(1\leq
i\leq d)$ of the output vector only depends on succeeding dimensions,
but not preceding ones. In other words, dimensions with larger
indices tend to end up getting little warping compared to the ones
with smaller indices. Fortunately, this can be easily resolved by a
\textit{Revert Layer}, which simply outputs a reversed version of its
input vector. Specifically, a Revert Layer $g$ operates as
\begin{align}
  g(\vect z):=g([z_1, z_2,...,z_d]^\top)=[z_d, z_{d-1},...,z_1]^\top
\end{align}
It's easy to verify a Revert Layer is bijective and that the Jacobian
of $g$ is a $d\times d$ matrix with 1s on its anti-diagonal and 0
otherwise, thus $\log\left|\det\frac{\partial g}{\partial \vect
  z}\right|$ is 0. Therefore, we can append a Revert Layer after each
ConvBlock to accommodate warping for dimensions with larger indices
without additional computation cost for the Jacobian as follows
\begin{align}
  \footnotesize
  \vect z\rightarrow \underbrace{\mbox{ConvBlock}\rightarrow \mbox{Revert}
    \rightarrow \mbox{ConvBlock}\rightarrow \mbox{Revert}\rightarrow... \rightarrow}_{\text{Repetitions of ConvBlock+Revert for }K\text{ times}} f(\vect z)
  \normalsize
\end{align}

\subsection{Connection to Inverse Autoregressive Flow}
\label{sec:compiaf}

Inspired by the idea of constructing complex tractable densities from
simpler ones with bijective transformations, different variants of the
original normalizing flow (NF)~\citep{DBLP:conf/icml/RezendeM15} have
been proposed. Perhaps the one most related to ConvFlow is Inverse
Autoregressive Flow~\citep{DBLP:conf/nips/KingmaSJCCSW16}, which employs
autoregressive transformations over the input dimensions to construct
output densities. Specifically, one layer of IAF works as follows
\begin{align}
  f(\vect z)=\vect\mu(\vect z)+\vect\sigma(\vect z)  \odot \vect z
  \label{eq:iaf}
\end{align}
where
\begin{align}
  [\vect\mu(\vect z), \vect\sigma(\vect z)]\leftarrow
  \text{AutoregressiveNN}(\vect z)
\end{align}
are outputs from an autoregressive neural network over the dimensions
of $\vect z$. There are two drawbacks of IAF compared to the proposed
ConvFlow:
\begin{itemize}
  \item The autoregressive neural network over input dimensions in
    IAF is represented by a Masked
    Autoencoder~\citep{germain2015made}, which generally requires
    $\mathcal{O}(d^2)$ parameters per layer, where $d$ is the input
    dimension, while each layer of ConvFlow is much more parameter
    efficient, only needing $k+d$ parameters ($k$ is the kernel size
    of 1d convolution and $k<d$).
  \item More importantly, due to the coupling of $\vect \sigma(\vect
    z)$ and $\vect z$ in the IAF transformation, in order to make the
    computation of the overall Jacobian determinant
    $\det\frac{\partial f}{\partial \vect z}$ linear in $d$, the
    Jacobian of the autoregressive NN transformation is assumed to be
    \textit{strictly} triangular (Equivalently, the Jacobian
    determinants of $\vect \mu$ and $\vect \sigma$ w.r.t $\vect z$ are
    both always 0. This is achieved by letting the $i$th dimension of
    $\vect \mu$ and $\vect \sigma$ depend only on dimensions
    $1,2,...,i-1$ of $\vect z$). In other words, \textit{the mappings
      from $\vect z$ onto $\vect \mu(\vect z)$ and $\vect \sigma(\vect
      z)$ via the autoregressive NN are always singular, no matter how
      their parameters are updated, and because of this, $\vect \mu$
      and $\vect \sigma$ will only be able to cover a subspace of the
      input space $\vect z$ belongs to}, which is obviously less
    desirable for a normalizing flow.\footnote{Since the singular
      transformations will only lead to subspace coverage of the
      resulting variable $\vect \mu$ and $\vect \sigma$, one could try
      to alleviate the subspace issue by modifying IAF to set both
      $\vect \mu$ and $\vect \sigma$ as free parameters to be learned,
      the resulting normalizing flow of which is exactly a version of
      planar flow as proposed in~\citep{DBLP:conf/icml/RezendeM15}.}
    Though these singularity transforms in the autoregressive NN are
    somewhat mitigated by their final coupling with the input $\vect
    z$, IAF still performs slightly worse in empirical evaluations
    than ConvFlow as no singular transform is involved in ConvFlow.
    \item Lastly, despite the similar nature of modeling variable
      dimension with an autoregressive manner, ConvFlow is much more
      efficient since the computation of the flow weights $w$ and the
      input $z$ is carried out by fast native 1-d convolutions, where
      IAF in its simplest form needs to maintain a masked feed forward
      network (if not maintaining an RNN). Similar idea of using
      convolution operators for efficient modeling of data dimensions
      is also adopted by PixelCNN~\cite{oord2016pixel}.
\end{itemize}

\section{Experiments}
\label{sec:exp}

We test performance the proposed ConvFlow on two settings, one on
synthetic data to infer unnormalized target density and the other on
density estimation for hand written digits and characters.

\subsection{Synthetic data}

\begin{figure}[!htb]
  \centering
  \begin{subfigure}[t]{0.15\textwidth}
    \includegraphics[width=\textwidth]{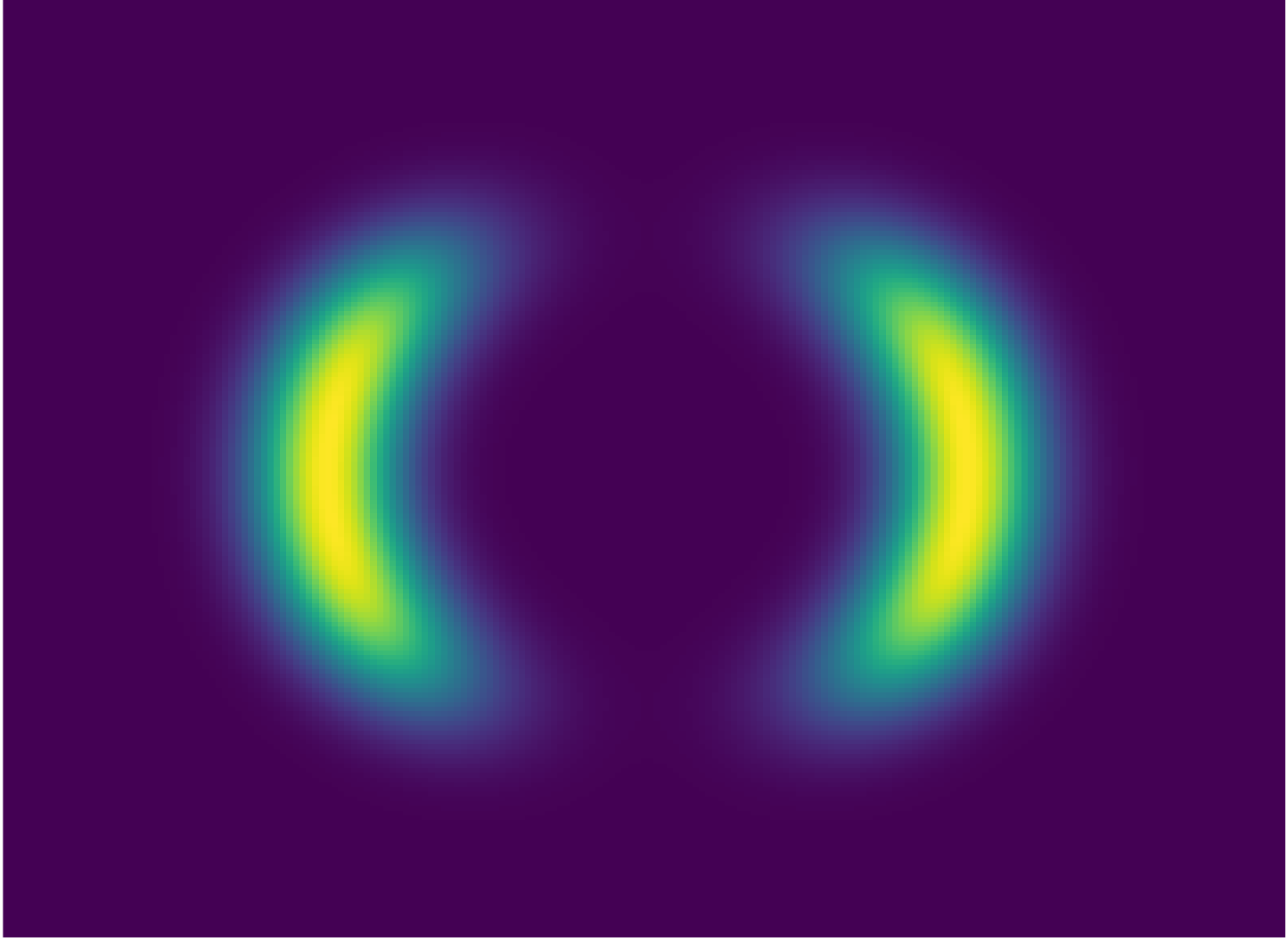}
  \end{subfigure}
  \begin{subfigure}[t]{0.15\textwidth}
    \includegraphics[width=\textwidth]{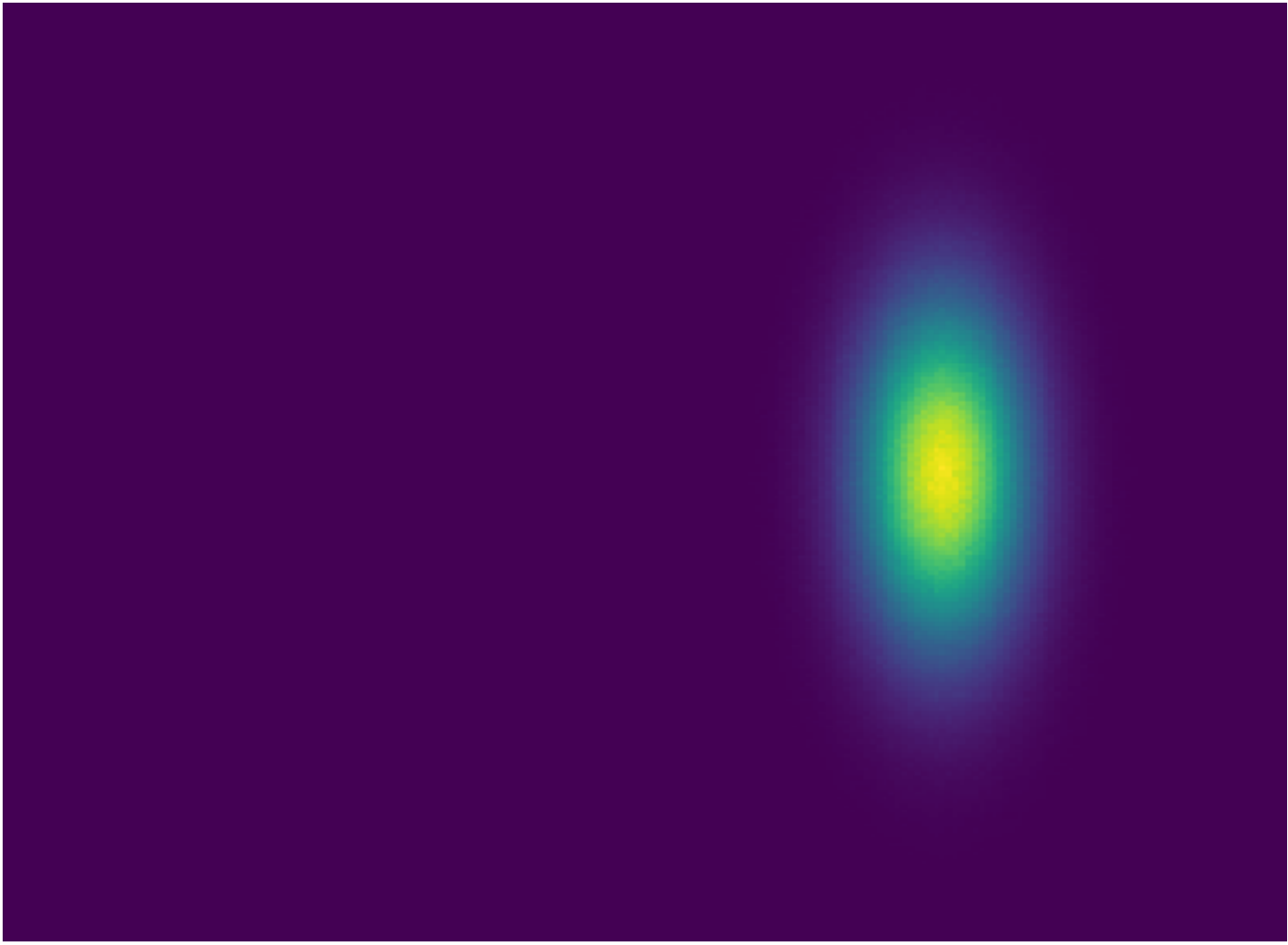}
  \end{subfigure}
  \begin{subfigure}[t]{0.15\textwidth}
    \includegraphics[width=\textwidth]{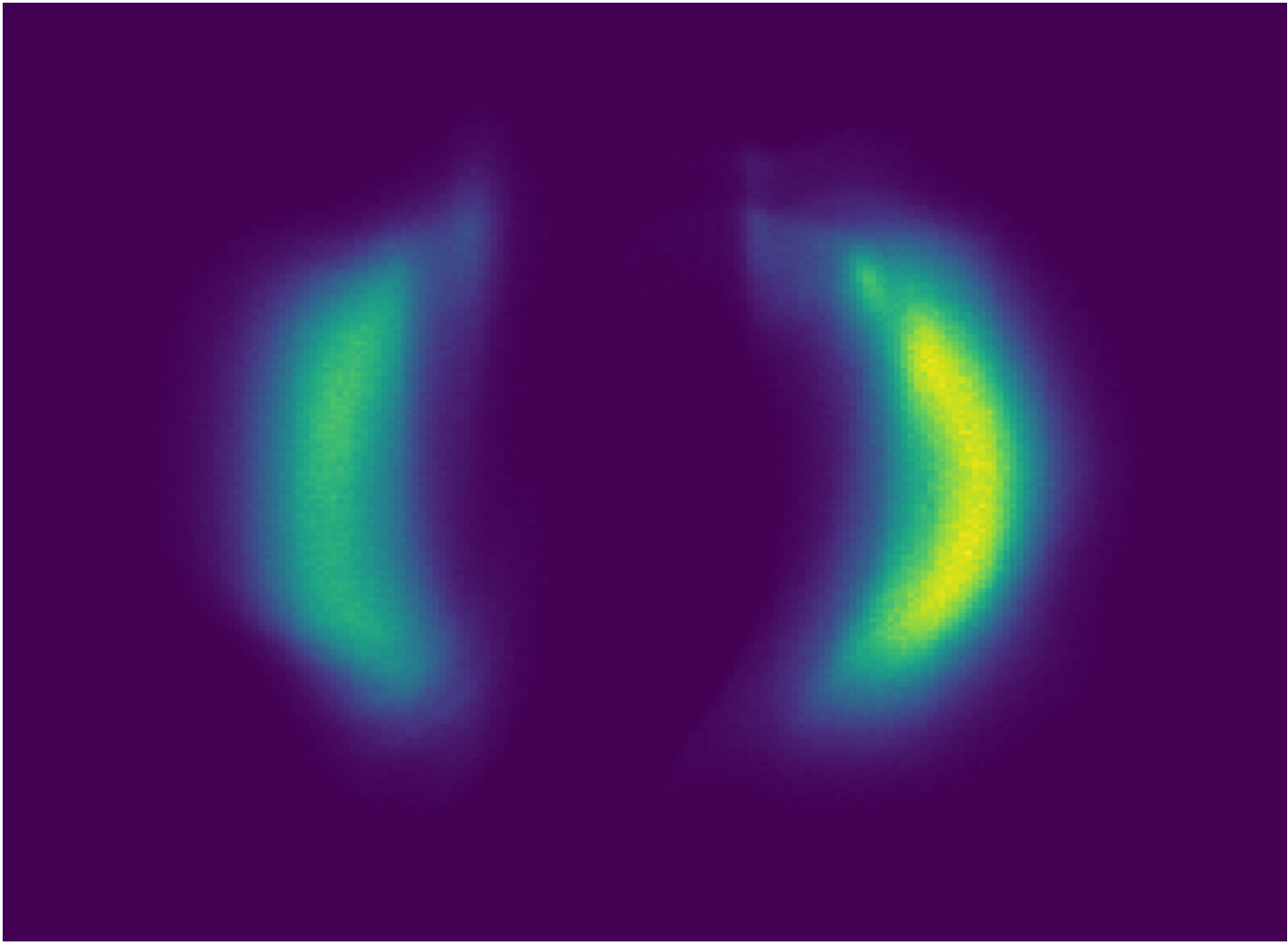}
  \end{subfigure}
  \\
  \begin{subfigure}[t]{0.15\textwidth}
      \includegraphics[width=\textwidth]{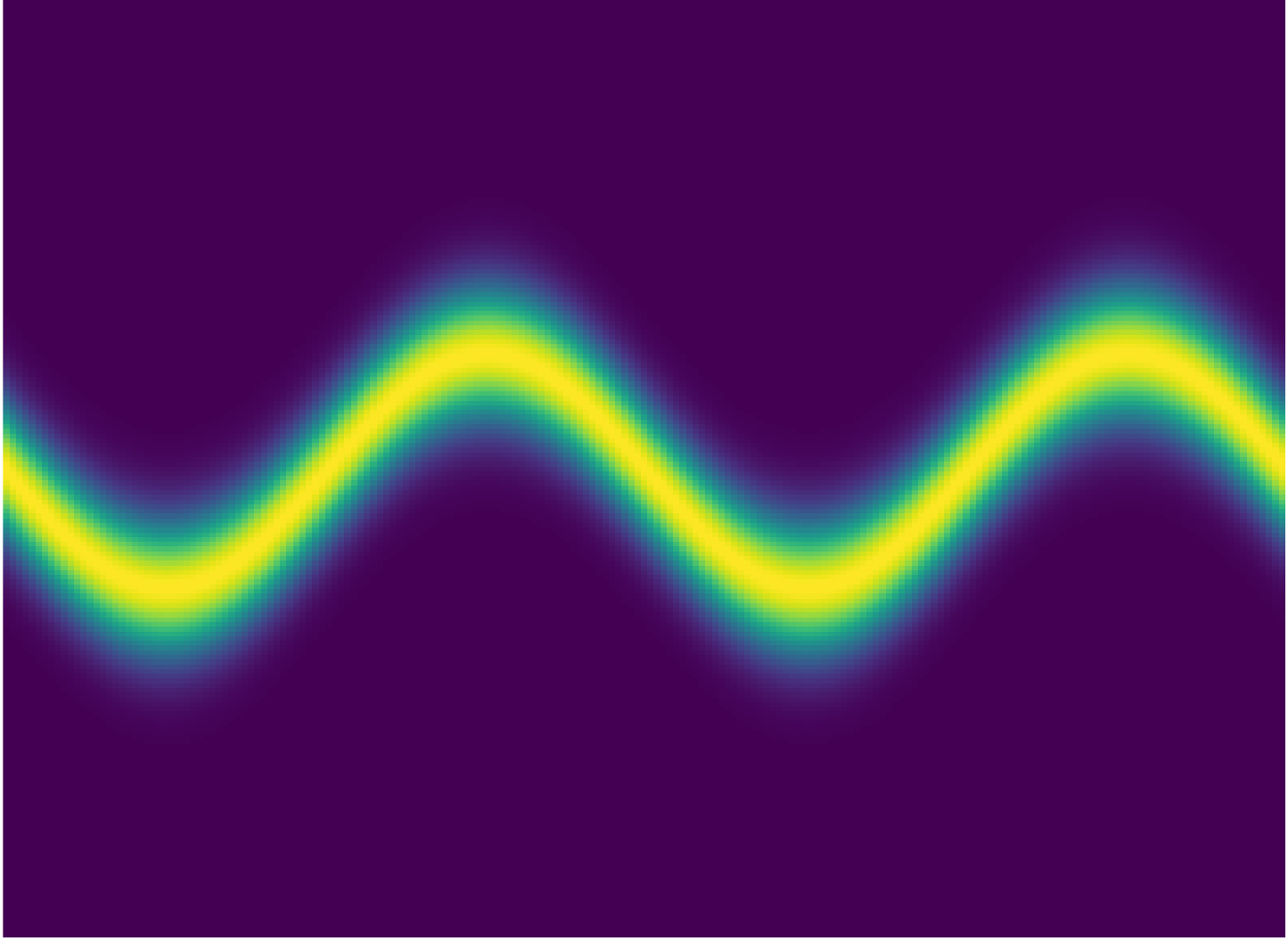}
    \end{subfigure}
  \begin{subfigure}[t]{0.15\textwidth}
    \includegraphics[width=\textwidth]{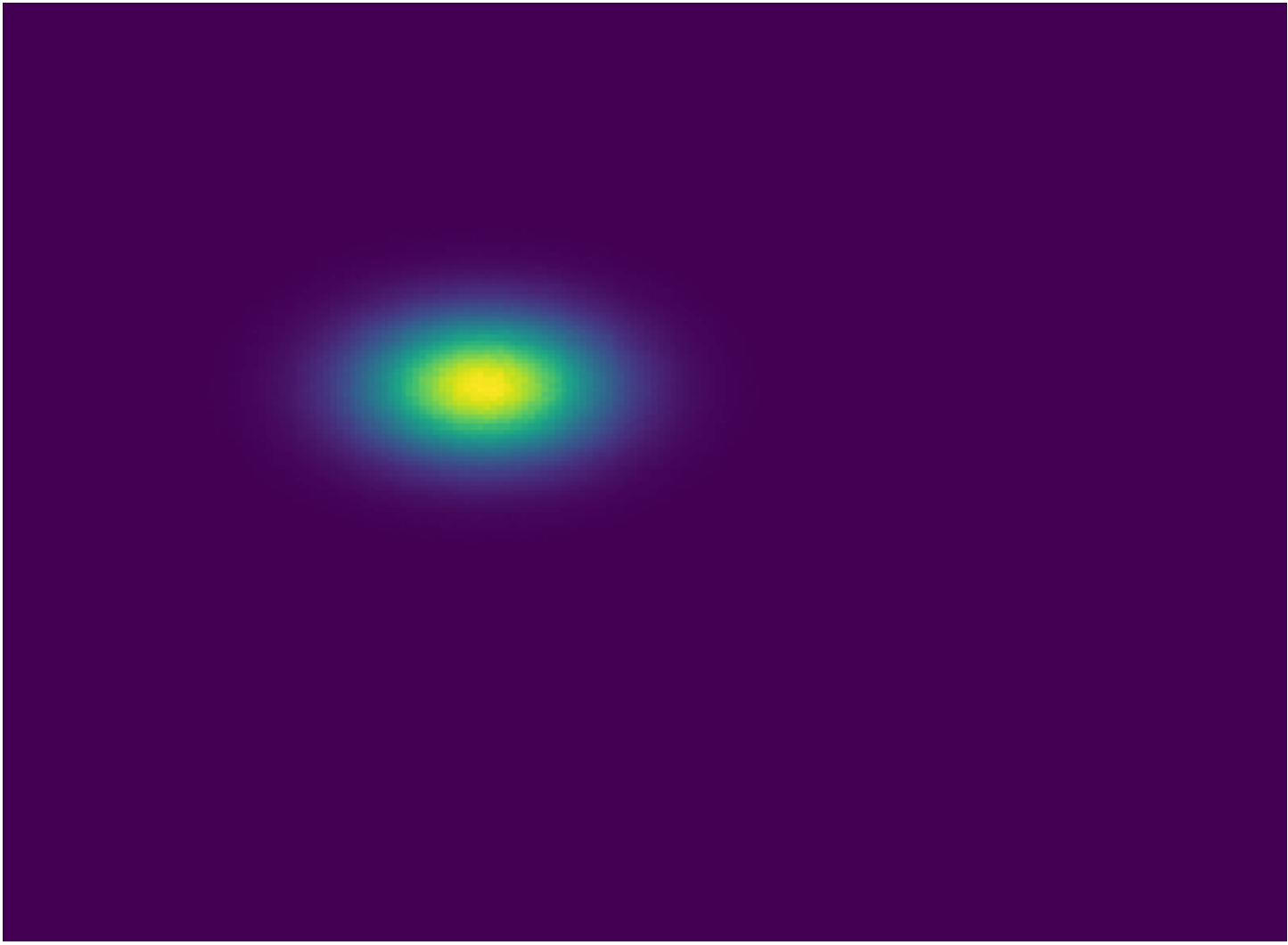}
    \label{fig:flow_vae}
  \end{subfigure}
  \begin{subfigure}[t]{0.15\textwidth}
    \includegraphics[width=\textwidth]{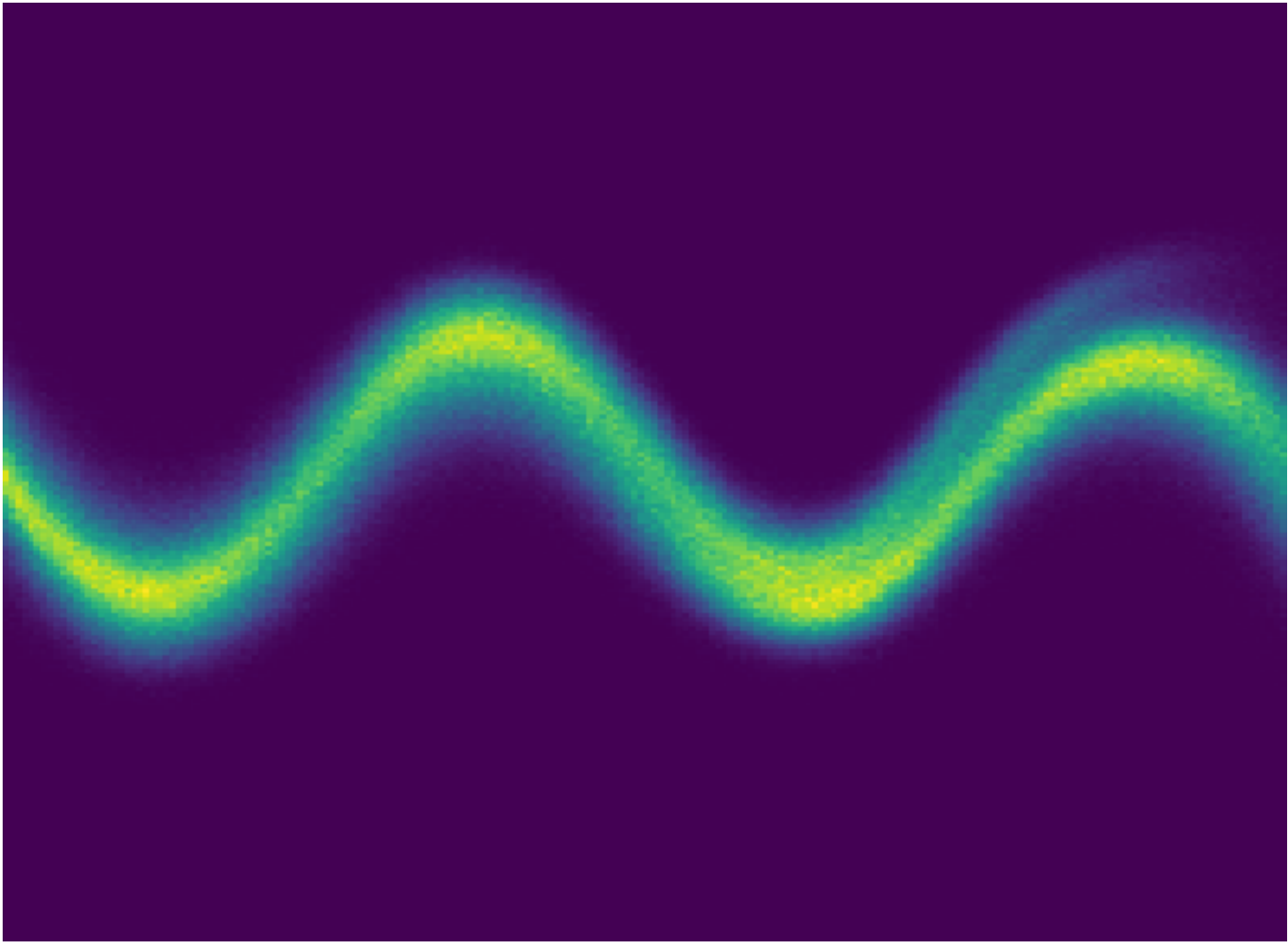}
    \label{fig:flow_avae}
  \end{subfigure}
  
  \caption{(a) True density; (b) Density learned by IAF (16 layers);
    (c) Density learned by ConvFlow. (8 blocks with each block
    consisting of 2 layers) }
  \label{fig:convdemo}
\end{figure}

We conduct experiments on using the proposed ConvFlow to approximate
an unnormalized target density of $\vect z$ with dimension 2 such that
$p(\vect z)\propto\exp(-U(\vect z))$. We adopt the same set of energy
functions $U(\vect z)$ in ~\cite{DBLP:conf/icml/RezendeM15} for a fair
comparison, which is reproduced below
\begin{align}
U_1(\vect z)&=\frac{1}{2}\left(\frac{\|\vect
  z\|-2}{4}\right)^2-\log\left(e^{-\frac{1}{2}\left[\frac{\vect
      z_1-2}{0.6}\right]^2}+e^{-\frac{1}{2}\left[\frac{\vect
      z_1+2}{0.6}\right]^2}\right)\nonumber\\
U_2(\vect z)&=\frac{1}{2}\left[\frac{\vect z_2-w_1(\vect z)}{0.4}\right]^2\nonumber
%U_3(\vect z)&=-\log\left(e^{-\frac{1}{2}\left[\frac{\vect
%      z_2-w_1(\vect z)}{0.35}\right]^2}+e^{-\frac{1}{2}\left[\frac{\vect
%      z_2-w_1(\vect z)+w_2(\vect z)}{0.35}\right]^2}\right)\nonumber\\
%U_4(\vect z)&=-\log\left(e^{-\frac{1}{2}\left[\frac{\vect
%      z_2-w_1(\vect z)}{0.4}\right]^2}+e^{-\frac{1}{2}\left[\frac{\vect
%      z_2-w_1(\vect z)+w_3(\vect z)}{0.35}\right]^2}\right)\nonumber
\end{align}
where $w_1(\vect z)=\sin\left(\frac{\pi\vect z_1}{2}\right)r$.
%w_2(\vect z)=3\exp\left(-\frac{1}{2}\left[\frac{\vect
%    z_2-1}{0.6}\right]^2\right), w_3(\vect
%z)=3\text{sigmoid}\left(\frac{\vect z_1-1}{0.3}\right)$.
The target
density of $z$ are plotted as the left most column in Figure
\ref{fig:convdemo}, and we test to see if the proposed ConvFlow can
transform a two dimensional standard Gaussian to the target density by
minimizing the KL divergence
\begin{align}
  & KL(q_K(\vect z_k)||p(\vect z))=\mathbb{E}_{\vect z_k}\log q_K(\vect z_k))-\mathbb{E}_{\vect z_k} \log p(\vect z_k)\nonumber\\
  =&\mathbb{E}_{\vect z_0}\log q_0(\vect z_0))-\mathbb{E}_{\vect z_0}\log\left|\det \frac{\partial f}{\partial z_0}\right|  +\mathbb{E}_{\vect z_0} U(f(\vect z_0))+\mbox{const}
\end{align}
where all expectations are evaluated with samples taken from
$q_0(z_0)$. We use a 2-d standard Gaussian as $q_0(z_0)$ and we test
different number of ConvBlocks stacked together in this task. Each
ConvBlock in this case consists a ConvFlow layer with kernel size 2,
dilation 1 and followed by another ConvFlow layer with kernel size 2,
dilation 2. Revert Layer is appended after each ConvBlock, and tanh
activation function is adopted by ConvFlow. The Autoregressive NN in
IAF is implemented as a two layer masked fully connected neural
network~\citep{germain2015made}.

Experimental results are shown in Figure \ref{fig:convdemo} for IAF
(middle column) and ConvFlow (right column) to approximate the target
density (left column). Even with 16 layers, IAF puts most of the
density to one mode, confirming our analysis about the singular
transform problem in IAF: As the data dimension is only two, the
subspace modeled by $\vect \mu(\vect z)$ and $\vect \sigma(\vect z)$
in Eq. (\ref{eq:iaf}) will be lying on a 1-d space, i.e., a straight
line, which is shown in the middle column. The effect of singular
transform on IAF will be less severe for higher dimensions. While with
8 layers of ConvBlocks (each block consists of 2 1d convolution
layers), ConvFlow is already approximating the target density quite
well despite the minor underestimate about the density around the
boundaries.

\subsection{Handwritten digits and characters}

\subsubsection{Setups}

To test the proposed ConvFlow for variational inference we use
standard benchmark datasets MNIST\footnote{Data downloaded from
  \url{http://www.cs.toronto.edu/~larocheh/public/datasets/binarized_mnist/}}
and OMNIGLOT\footnote{Data downloaded from
  \url{https://github.com/yburda/iwae/raw/master/datasets/OMNIGLOT/chardata.mat}}~\citep{DBLP:conf/nips/LakeST13}. Our
method is general and can be applied to any formulation of the
generative model $p_\theta(x,z)$; For simplicity and fair comparison,
in this paper, we focus on densities defined by stochastic neural
networks, i.e., a broad family of flexible probabilistic generative
models with its parameters defined by neural networks. Specifically,
we consider the following two family of generative models
\begin{align}
  &\textsc{G}_1: p_\theta(x,z)=p_\theta(z)p_\theta(x|z)\\
  &\textsc{G}_2: p_\theta(x,z_1, z_2)=p_\theta(z_1)p_\theta(z_2|z_1)p_\theta(x|z_2)
\end{align}
where $p(z)$ and $p(z_1)$ are the priors defined over $z$ and $z_1$
for $G_1$ and $G_2$, respectively. All other conditional densities are
specified with their parameters $\theta$ defined by neural networks,
therefore ending up with two stochastic neural networks. This network
could have any number of layers, however in this paper, we focus on the
ones which only have one and two stochastic layers, i.e., $G_1$ and
$G_2$, to conduct a fair comparison with previous methods on similar
network architectures, such as VAE, IWAE and Normalizing Flows.

We use the same network architectures
for both $G_1$ and $G_2$ as in~\citep{DBLP:journals/corr/BurdaGS15},
specifically shown as follows
\begin{enumerate}
\item[$G_1:$] A single Gaussian stochastic layer $z$ with 50 units. In between the
  latent variable $z$ and observation $x$ there are two deterministic
  layers, each with 200 units;
\item[$G_2:$] Two Gaussian stochastic layers $z_1$ and $z_2$ with 50 and 100
  units, respectively. Two deterministic layers with 200 units connect
  the observation $x$ and latent variable $z_2$, and two deterministic
  layers with 100 units are in between $z_2$ and $z_1$.
\end{enumerate}
where a Gaussian stochastic layer consists of two fully connected
linear layers, with one outputting the mean and the other outputting the
logarithm of diagonal covariance. All other deterministic layers are
fully connected with tanh nonlinearity.
\begin{comment}
For $G_1$, inference network with the following architecture is used
\begin{align}
  \tau_1&=f_1(x\|\epsilon_1)\,\mbox{ where }\epsilon_1\sim\mathcal{N}(0, I)\\
  \tau_{i}&=f_i(\tau_{i-1}\|\epsilon_i)\,\mbox{ where }\epsilon_i\sim\mathcal{N}(0, I)\nonumber\\ &\mbox{ for $i=2,...,k$}\\
  q(z|x, \tau_1,...,\tau_k)&=\mathcal{N}\Big(\mu(x\|\tau_1\|...\|\tau_k),\mbox{diag}\big(\sigma(x\|\tau_1\|...\|\tau_k)\Big)
\end{align}
where $\|$ denotes the concatenation operator. All noise vectors
$\epsilon$s are set to be of 50 dimensions, and all other variables
have the corresponding dimensions in the generative model. Inference
network used for $G_2$ is the same, except for the Gaussian stochastic
layer is defined for $z_2$. An additional Gaussian stochastic layer
with $z_2$ as input is defined for $z_1$ with the dimensions of
variables aligned to those in the generative model $G_2$.  Further,
\end{comment}
Bernoulli observation models are assumed for both MNIST and
OMNIGLOT. For MNIST, we employ the static binarization strategy as
in~\citep{DBLP:journals/jmlr/LarochelleM11} while dynamic binarization
is employed for OMNIGLOT.

The inference networks $q(z|x)$ for $G_1$ and $G_2$ have similar
architectures to the generative models, with details
in~\citep{DBLP:journals/corr/BurdaGS15}. ConvFlow is hence used to
warp the output of the inference network $q(z|x)$, assumed be to
Gaussian conditioned on the input $x$, to match complex true
posteriors. Our baseline models include
VAE~\citep{DBLP:journals/corr/KingmaW13},
IWAE~\citep{DBLP:journals/corr/BurdaGS15} and Normalizing
Flows~\citep{DBLP:conf/icml/RezendeM15}. Since our propose method
involves adding more layers to the inference network, we also include
another enhanced version of VAE with more deterministic layers added
to its inference network, which we term as VAE+.\footnote{VAE+ adds
  more layers before the stochastic layer of the inference network
  while the proposed method is add convolutional flow layers after the
  stochastic layer.} With the same VAE architectures, we also test the
abilities of constructing complex variational posteriors with IAF and
ConvFlow, respectively. All models are implemented in
PyTorch. Parameters of both the variational distribution and the
generative distribution of all models are optimized with
Adam~\citep{DBLP:journals/corr/KingmaB14} for 2000 epochs, with a
fixed learning rate of 0.0005, exponential decay rates for the 1st and
2nd moments at 0.9 and 0.999, respectively. Batch
normalization~\citep{DBLP:conf/icml/IoffeS15} and linear annealing of
the KL divergence term between the variational posterior and the prior
is employed for the first 200 epochs, as it has been shown to help
training multi-layer stochastic neural
networks~\cite{DBLP:conf/nips/SonderbyRMSW16}. Code to reproduce all
reported results will be made publicly available.%For ASY-VAE-ER and
%ASY-IWAE-ER, the entropy regularization parameter $\lambda$ are tuned
%with cross validation on the training set, from a range of $\{10^{-1},
%10^{-2}, 10^{-3}, 10^{-4}, 10^{-5}\}$.

For inference models with latent variable $\vect z$ of 50 dimensions,
a ConvBlock consists of following ConvFlow layers
\begin{align}
  [&\text{ConvFlow}(\text{kernel size}=5 \text{, dilation}=1),\nonumber\\
    &\text{ConvFlow}(\text{kernel size}=5 \text{, dilation}=2),\nonumber\\
    &\text{ConvFlow}(\text{kernel size}=5 \text{, dilation}=4),\nonumber\\
    &\text{ConvFlow}(\text{kernel size}=5 \text{, dilation}=8),\nonumber\\
    &\text{ConvFlow}(\text{kernel size}=5 \text{, dilation}=16),\nonumber\\
    &\text{ConvFlow}(\text{kernel size}=5 \text{, dilation}=32)]
\end{align}
and for inference models with latent variable $\vect z$ of 100
dimensions, a ConvBlock consists of following ConvFlow layers
\begin{align}
    [&\text{ConvFlow}(\text{kernel size}=5 \text{, dilation}=1),\nonumber\\
    &\text{ConvFlow}(\text{kernel size}=5 \text{, dilation}=2),\nonumber\\
    &\text{ConvFlow}(\text{kernel size}=5 \text{, dilation}=4),\nonumber\\
    &\text{ConvFlow}(\text{kernel size}=5 \text{, dilation}=8),\nonumber\\
    &\text{ConvFlow}(\text{kernel size}=5 \text{, dilation}=16),\nonumber\\
    &\text{ConvFlow}(\text{kernel size}=5 \text{, dilation}=32),\nonumber\\
    &\text{ConvFlow}(\text{kernel size}=5 \text{, dilation}=64)]
\end{align}
A Revert layer is appended after each ConvBlock and leaky ReLU with a
negative slope of $0.01$ is used as the activation function in
ConvFlow. For IAF, the autoregressive neural network is implemented as
a two layer masked fully connected neural network.

\subsubsection{Generative Density Estimation}

For MNIST, models are trained and tuned on the 60,000 training and
validation images, and estimated log-likelihood on the test set with
128 importance weighted samples are reported. Table \ref{tab:mnist}
presents the performance of all models, when the generative model is
assumed to be from both $G_1$ and $G_2$.

Firstly, VAE+ achieves higher log-likelihood estimates than vanilla
VAE due to the added more layers in the inference network, implying
that a better posterior approximation is learned (which is still
assumed to be a Gaussian). Second, we observe that VAE with ConvFlow
achieves much better density estimates than VAE+, which confirms our
expectation that warping the variational distribution with
convolutional flows enforces the resulting variational posterior to
match the true non-Gaussian posterior. Also, adding more blocks of
convolutional flows to the network makes the variational posterior
further close to the true posterior. We also observe that VAE with
Inverse Autoregressive Flows (VAE+IAF) improves over VAE and VAE+, due
to its modeling of complex densities, however the improvements are not
as significant as ConvFlow. The limited improvement might be explained
by our analysis on the singular transformation and subspace issue in
IAF. Lastly, combining convolutional normalizing flows with multiple
importance weighted samples, as shown in last row of Table
\ref{tab:mnist}, further improvement on the test set log-likelihood is
achieved. Overall, the method combining ConvFlow and importance
weighted samples achieves best NLL on both settings, outperforming
IWAE significantly by 7.1 nats on $G_1$ and 5.7 nats on $G_2$. Notice
that, ConvFlow combined with IWAE achieves an NLL of 79.11, comparable
to the best published result of 79.10, achieved by
PixelRNN~\citep{oord2016pixel} with a much more sophisticated
architecture. Also it's about 0.8 nat better than the best IAF result
of 79.88 reported in~\citep{DBLP:conf/nips/KingmaSJCCSW16}, which
demonstrates the representative power of ConvFlow compared to
IAF\footnote{The result in ~\citep{DBLP:conf/nips/KingmaSJCCSW16} are
  not directly comparable, as their results are achieved with a much
  more sophisticated VAE architecture and a much higher dimension of
  latent code ($d=1920$ for the best NLL of 79.88). However, in this
  paper, we only assume a relatively simple VAE architecture compose
  of fully connected layers and the dimension of latent codes to be
  relatively low, 50 or 100, depending on the generative model in
  VAE. One could expect the performance of ConvFlow to improve even
  further if similar complex VAE architecture and higher dimension of
  latent codes are used.}.

\begin{table*}[htb]
  \centering
  \caption{MNIST test set NLL with generative models $G_1$ and $G_2$ (lower is
    better $K$ is number of ConvBlocks)}
  \begin{tabular}{lrr}
    \toprule
    MNIST (static binarization) & $-\log p(x)$ on $G_1$  & $-\log p(x)$ on $G_2$ \\
    \midrule
    \textsf{VAE} ~\citep{DBLP:journals/corr/BurdaGS15}       & $88.37$ & $85.66$\\ %new
    %    IWAE $(IW=50)$~\citep{DBLP:journals/corr/BurdaGS15} & \tablefootnote{This is the result from our own implementation of IWAE, which is better than the original result of $87.10$ reported in ~\citep{DBLP:journals/corr/BurdaGS15} for the same network configuration.}$87.10(\textcolor{cyan}{85.65})$ & $85.32 (84.04)$\\
    \textsf{IWAE} $(IW=50)$~\citep{DBLP:journals/corr/BurdaGS15} & $86.90$ & $84.26$\\
    \textsf{VAE+NF}~\citep{DBLP:conf/icml/RezendeM15} & - & $\leq 85.10$\\
    \midrule
    \textsf{VAE+} $(K=1)$ & $88.20$ & $ 85.41$ \\ %new
    \textsf{VAE+} $(K=4)$ & $88.08$ & $ 85.26$ \\ %new
    \textsf{VAE+} $(K=8)$ & $87.98$ & $ 85.16$ \\ %new
    \midrule
    \textsf{VAE+IAF} $(K=1)$ & $87.70$ & $ 85.03$ \\ %new
    \textsf{VAE+IAF} $(K=2)$ & $87.30$ & $ 84.74$ \\ %new
    \textsf{VAE+IAF} $(K=4)$ & $87.02$ & $ 84.55$ \\ %new
    \textsf{VAE+IAF} $(K=8)$ & $86.62$ & $ 84.26$ \\ %new
    \midrule
    \textsf{VAE+ConvFlow} $(K=1)$ & $86.91$ & $ {85.45}$ \\ %old
    \textsf{VAE+ConvFlow} $(K=2)$ & $86.40$ & $ {85.37}$ \\ %old
    \textsf{VAE+ConvFlow} $(K=4)$ & $84.78$ & $ {81.64}$ \\ %old
    \textsf{VAE+ConvFlow} $(K=8)$ & $83.89$ & $ {81.21}$ \\ %old
    \midrule
    \textsf{IWAE+ConvFlow} $(K=8, IW=50)$ & $79.78$& $79.11$\\%sim
%    ASY-IWAE $(IW=50, k=1)$ & $85.76 $ & $ {83.77} $\\
%    ASY-IWAE $(IW=50, k=4)$ & $85.31 $ & $ {83.52} $\\
%    ASY-IWAE $(IW=50, k=8)$ & $85.03 $ & $ {83.29} $\\
    \bottomrule
  \end{tabular}
  \label{tab:mnist}
\end{table*}

\begin{table*}[htb]
  \centering
  \caption{OMNIGLOT test set NLL with generative models $G_1$ and $G_2$ (lower is better, $K$ is number of ConvBlocks)}
  \begin{tabular}{lrr}
    \toprule
    OMNIGLOT  & $-\log p(x)$ on $G_1$ & $-\log p(x)$ on $G_2$ \\
    \midrule
    \textsf{VAE}~\citep{DBLP:journals/corr/BurdaGS15}       & $108.22$ & $106.09$ \\
    \textsf{IWAE} $(IW=50)$~\citep{DBLP:journals/corr/BurdaGS15} & $106.08$ & $104.14$ \\
    \midrule
    \textsf{VAE+} $(K=1)$ & $108.30$ & $ {106.30}$ \\
    \textsf{VAE+} $(K=4)$ & $108.31$ & $ {106.48}$ \\ 
    \textsf{VAE+} $(K=8)$ & $108.31$ & $ {106.05}$ \\
    \midrule
    \textsf{VAE+IAF} $(K=1)$ & $107.31$ & $ 105.78$ \\
    \textsf{VAE+IAF} $(K=2)$ & $106.93$ & $ 105.34$ \\
    \textsf{VAE+IAF} $(K=4)$ & $106.69$ & $ 105.56$ \\ 
    \textsf{VAE+IAF} $(K=8)$ & $106.33$ & $ 105.00$ \\
    \midrule
    \textsf{VAE+ConvFlow} $(K=1)$ & $106.42$ & $ 105.33$ \\% sim
    \textsf{VAE+ConvFlow} $(K=2)$ & $106.08$ & $ 104.85$ \\% sim
    \textsf{VAE+ConvFlow} $(K=4)$ & $105.21$ & $ 104.30$ \\% sim
    \textsf{VAE+ConvFlow} $(K=8)$ & $104.86$ & $ 103.49$ \\% sim
%    ASY-IWAE $(IW=50, k=1)$ & $104.83 $ & $ {103.57} $ \\% sim
%    ASY-IWAE $(IW=50, k=4)$ & $104.80 $ & $ {103.44}$ \\% sim
    %    ASY-IWAE $(IW=50, k=8)$ & $104.63 $ & $ {103.40} $ \\% sim
    \midrule
    \textsf{IWAE+ConvFlow} $(K=8, IW=50)$ & $104.21$ & $ 103.02$ \\% sim
    \bottomrule
  \end{tabular}
  \label{tab:omniglot}
\end{table*}

Results on OMNIGLOT are presented in Table \ref{tab:omniglot} where
similar trends can be observed as on MNIST. One observation different
from MNIST is that, the gain from IWAE+ConvFlow over IWAE is not as
large as it is on MNIST, which could be explained by the fact that
OMNIGLOT is a more difficult set compared to MNIST, as there are 1600
different types of symbols in the dataset with roughly 20 samples per
type. Again on OMNIGLOT we observe IAF with VAE improves over VAE and
VAE+, while doesn't perform as well as ConvFlow.

\subsubsection{Latent Code Visualization}
\begin{figure*}[!htb]
  \centering \includegraphics[height=0.23\linewidth]{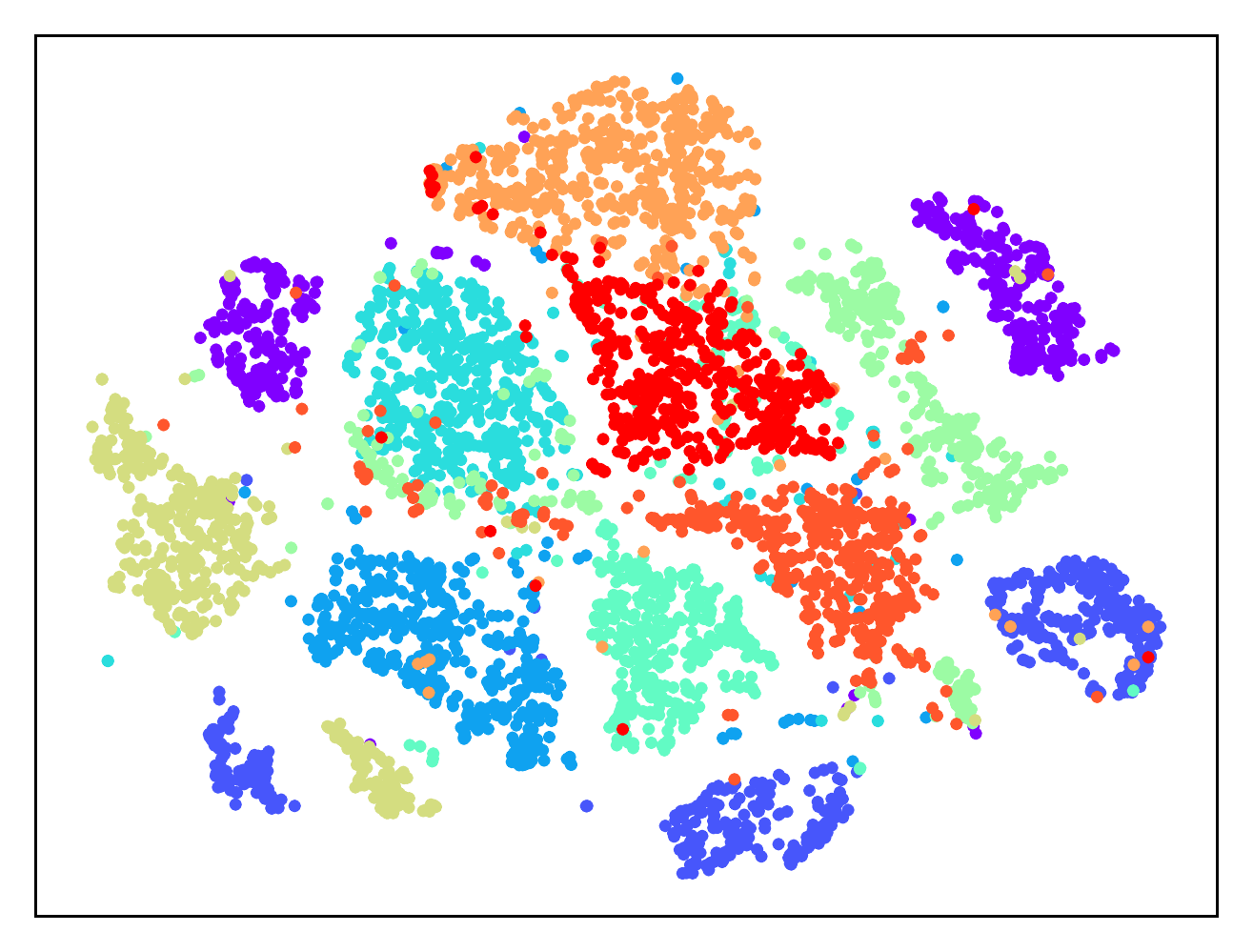}
  \includegraphics[height=0.23\linewidth]{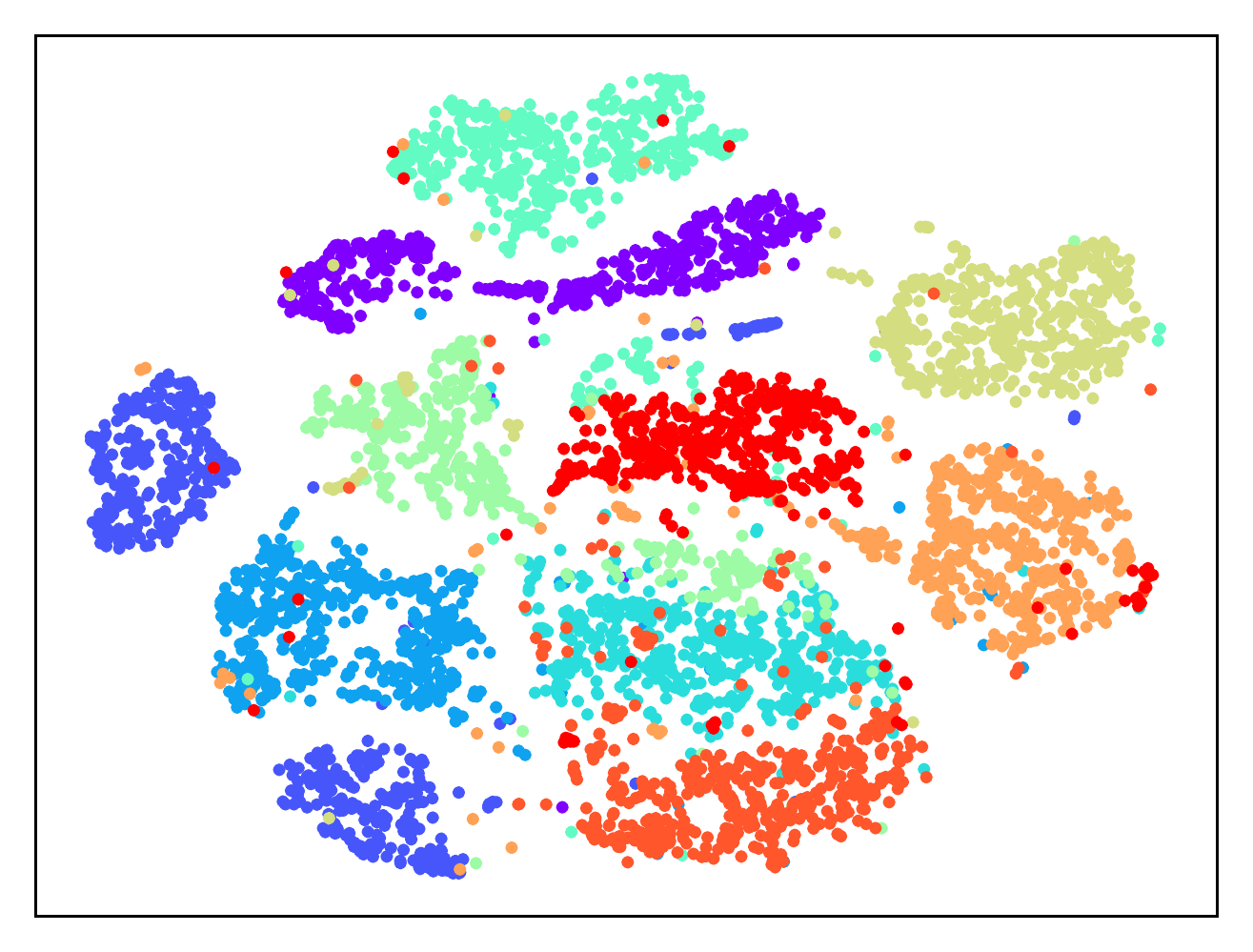}
  \includegraphics[height=0.23\linewidth]{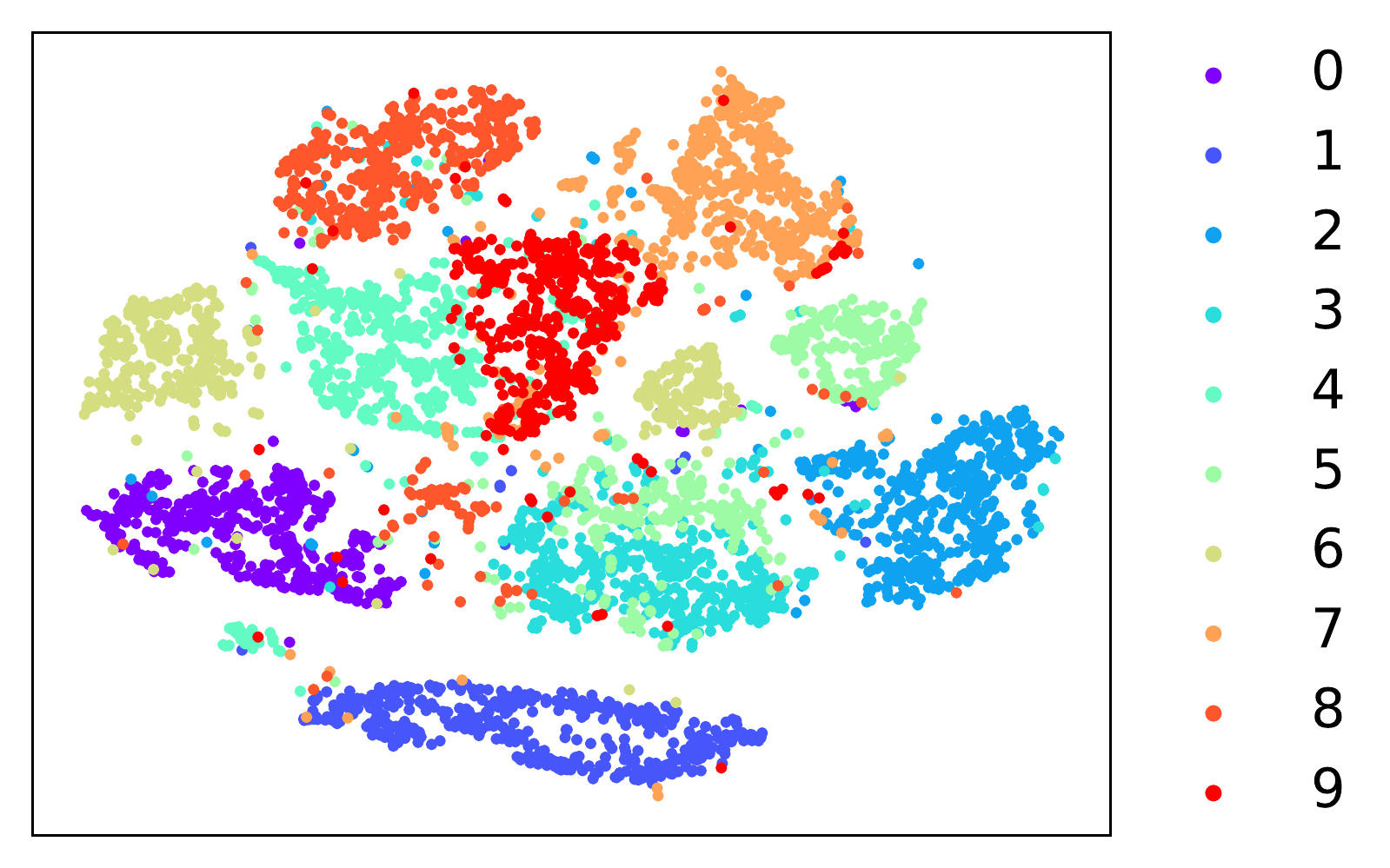}
  \caption{\textbf{Left:} VAE, \textbf{Middle:} VAE+IAF,
    \textbf{Right:}VAE+ConvFlow. (best viewed in color)}
  \label{fig:tsne}
\end{figure*}

We visualize the inferred latent codes $z$ of 5000 digits in the MNIST
test set with respect to their true class labels in Figure
\ref{fig:tsne} from different models with
tSNE~\cite{maaten2008visualizing}.
We observe that on generative model $G_2$, all three models are able
to infer latent codes of the digits consistent with their true
classes. However, VAE and VAE+IAF both show disconnected cluster of
latent codes from the same class (e.g., digits 0 and digits 1). Latent
codes inferred by VAE for digit 3 and 5 tend to mix with each other.
Overall, VAE equipped with ConvFlow produces clear separable latent
codes for different classes while also maintaining high in-class
density (notably for digit classes 0, 1, 2, 7, 8, 9 as shown in the
rightmost figure).

\subsubsection{Generation}

\begin{figure*}[!htb]
  \centering
  \begin{subfigure}[t]{0.245\textwidth}
    \includegraphics[width=\textwidth]{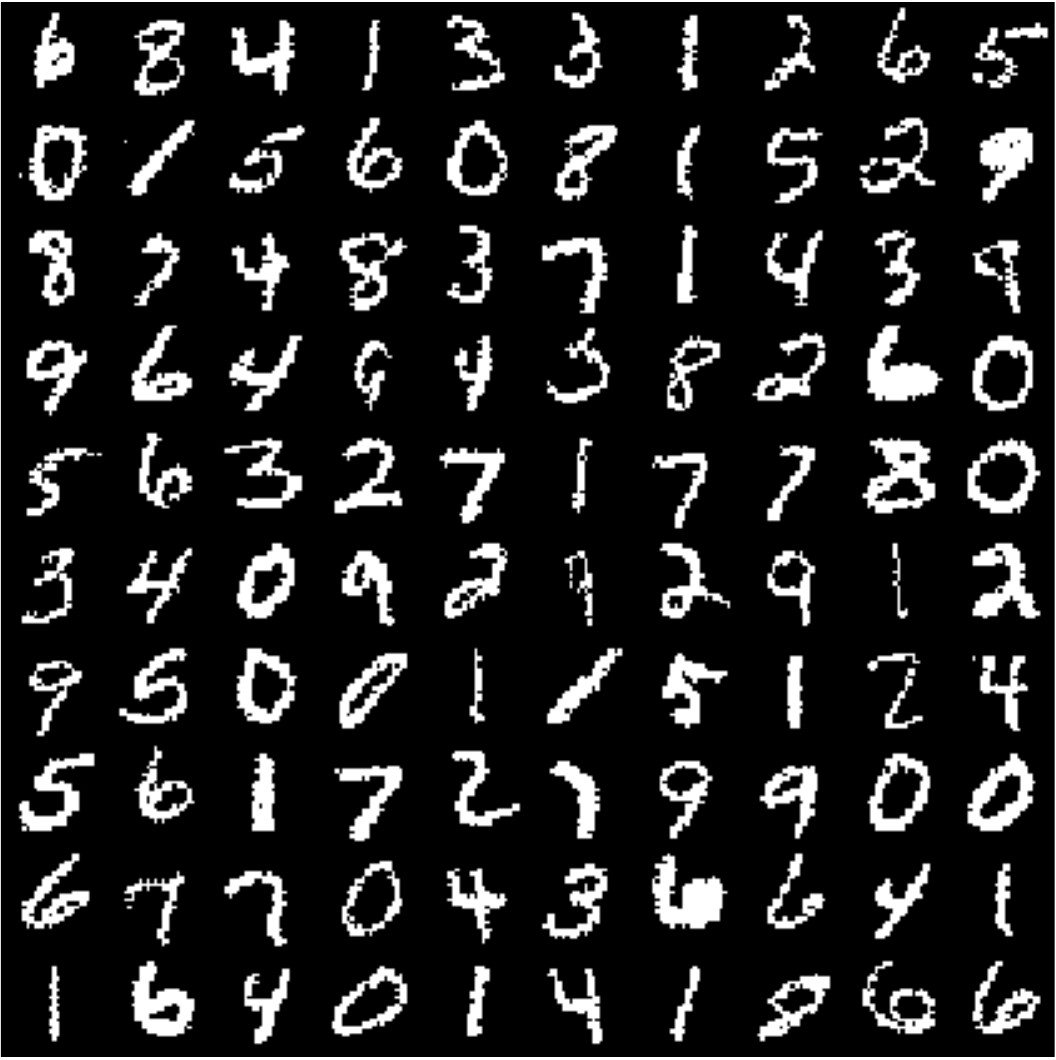}
    \caption{MNIST Training data}
  \end{subfigure}
  \begin{subfigure}[t]{0.245\textwidth}
    \includegraphics[width=\textwidth]{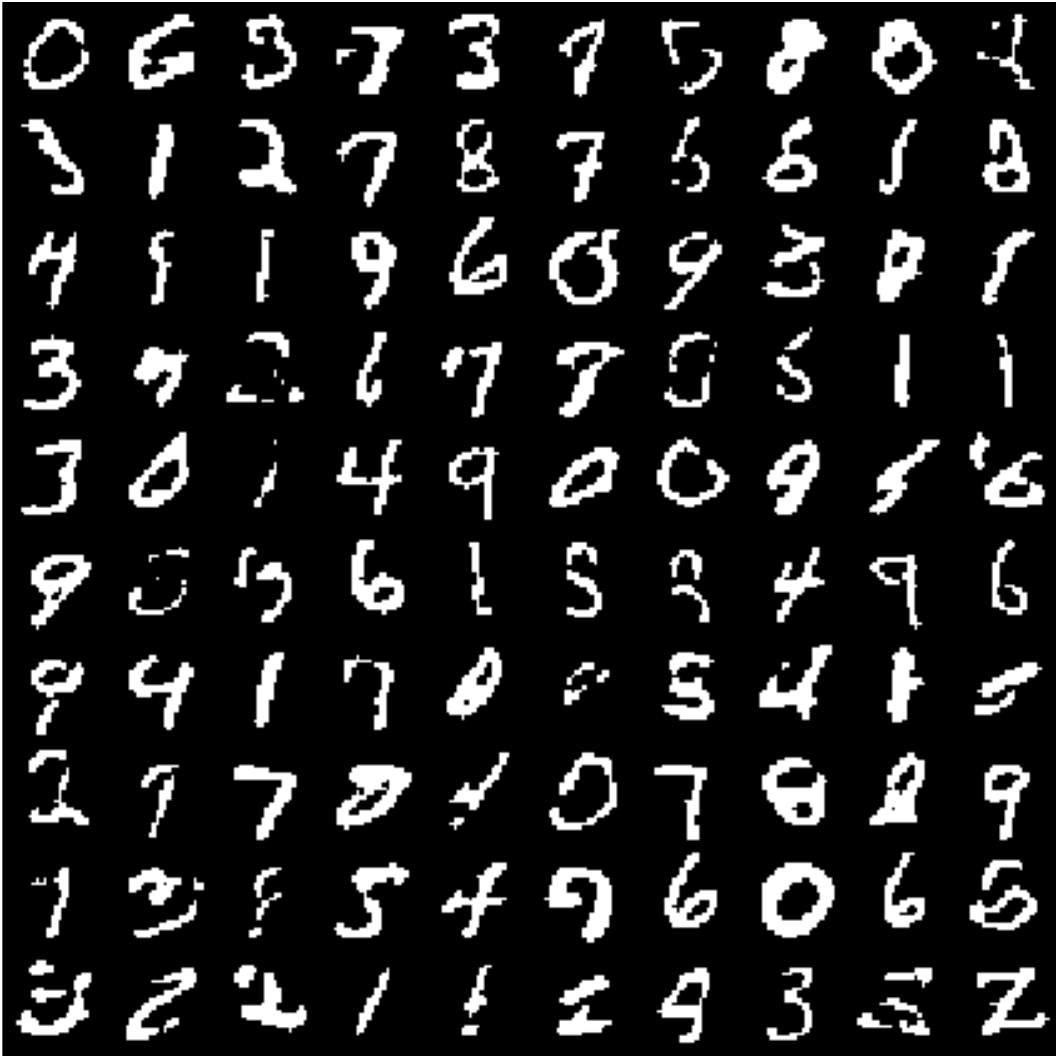}
    \caption{Random samples 1 from IWAE-ConvFlow ($K=8$)}
  \end{subfigure}
  \begin{subfigure}[t]{0.245\textwidth}
    \includegraphics[width=\textwidth]{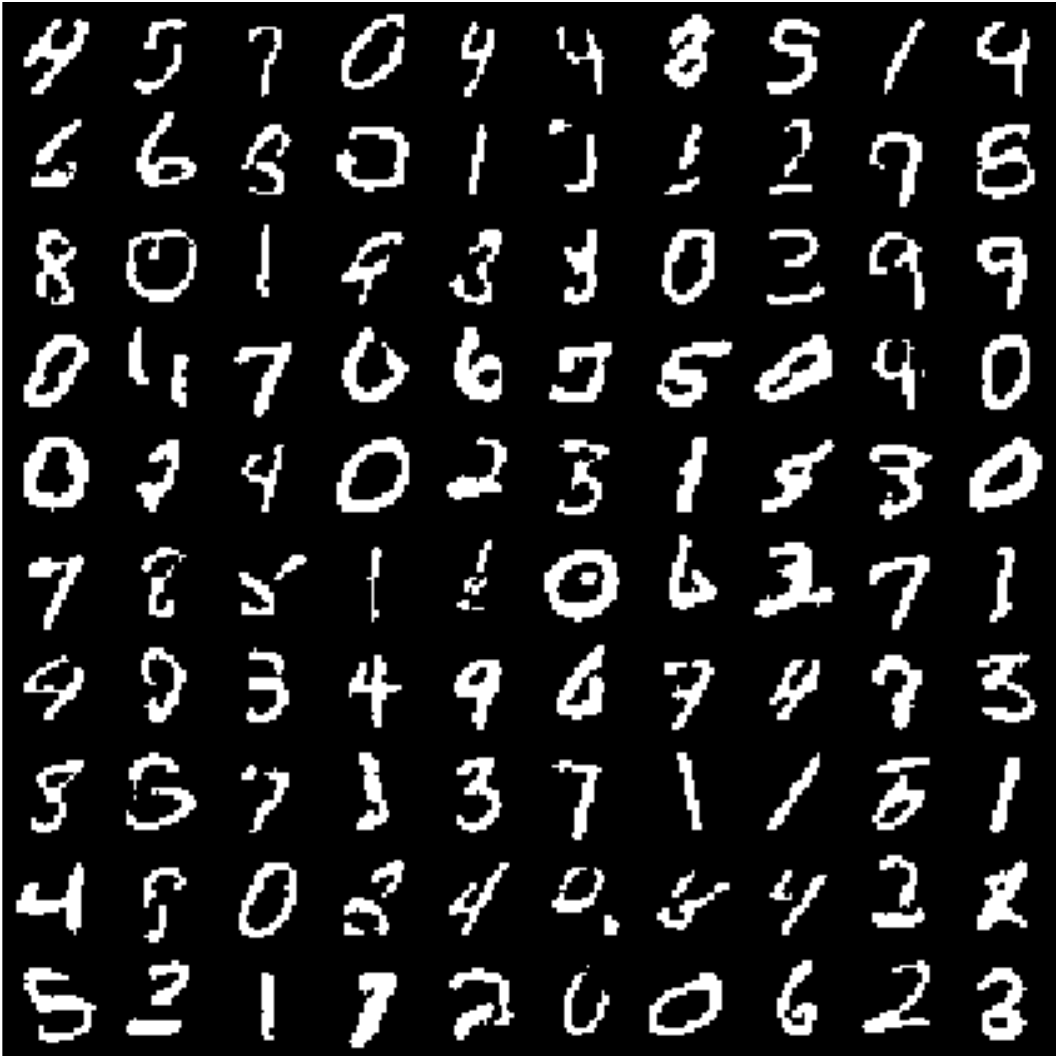}
    \caption{Random samples 2 from IWAE-ConvFlow ($K=8$)}
  \end{subfigure}
  \begin{subfigure}[t]{0.245\textwidth}
    \includegraphics[width=\textwidth]{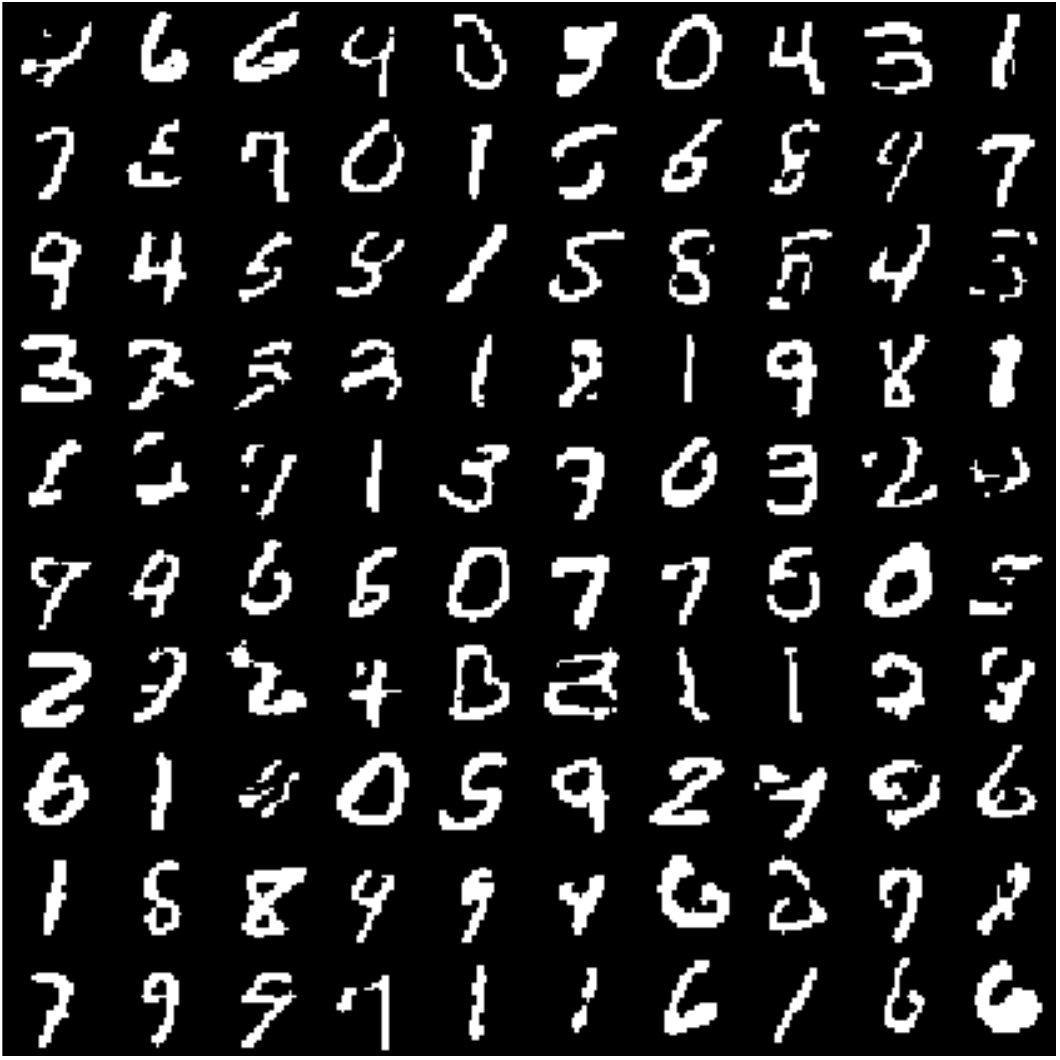}
    \caption{Random samples 3 from IWAE-ConvFlow ($K=8$)}
  \end{subfigure}
  \begin{subfigure}[t]{0.245\textwidth}
    \includegraphics[width=\textwidth]{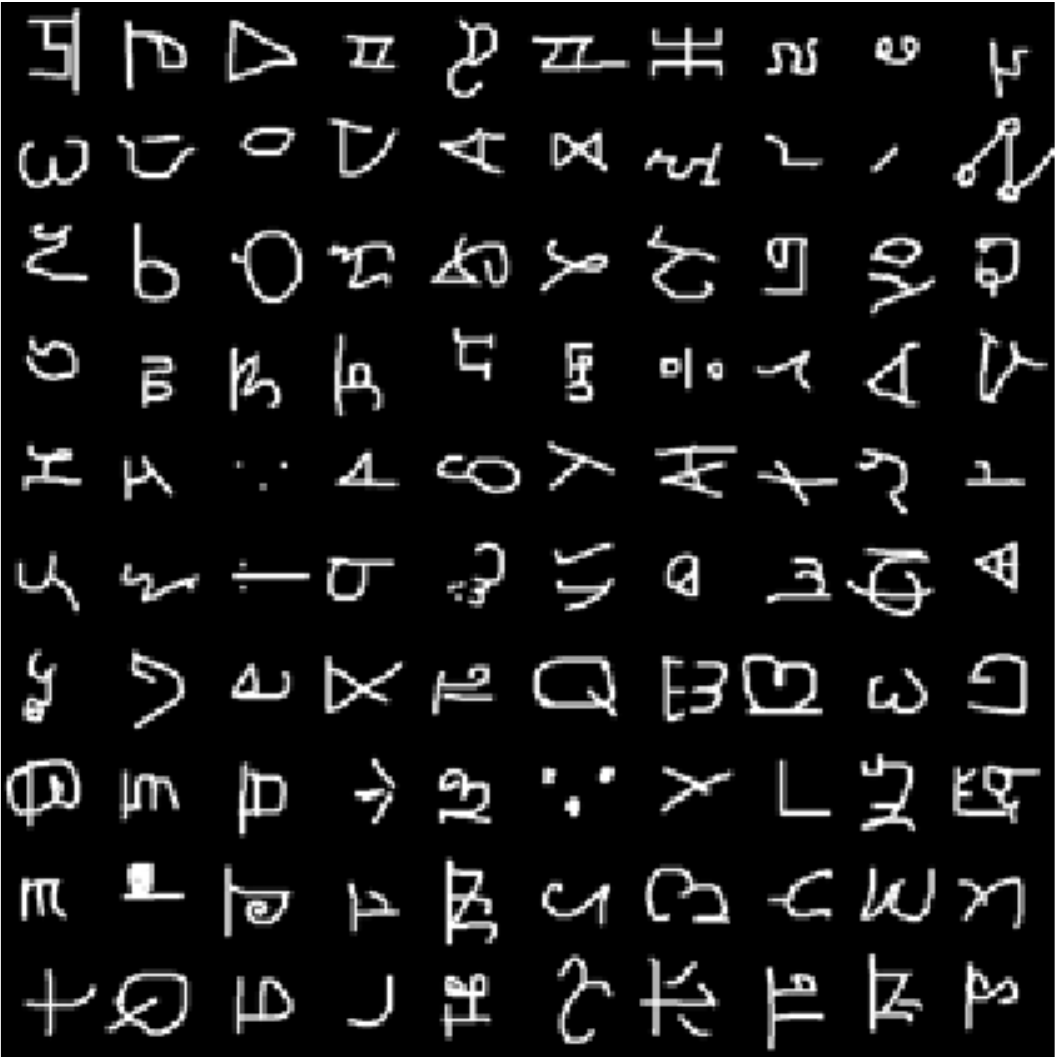}
    \caption{OMNIGLOT Training data}
  \end{subfigure}
  \begin{subfigure}[t]{0.245\textwidth}
    \includegraphics[width=\textwidth]{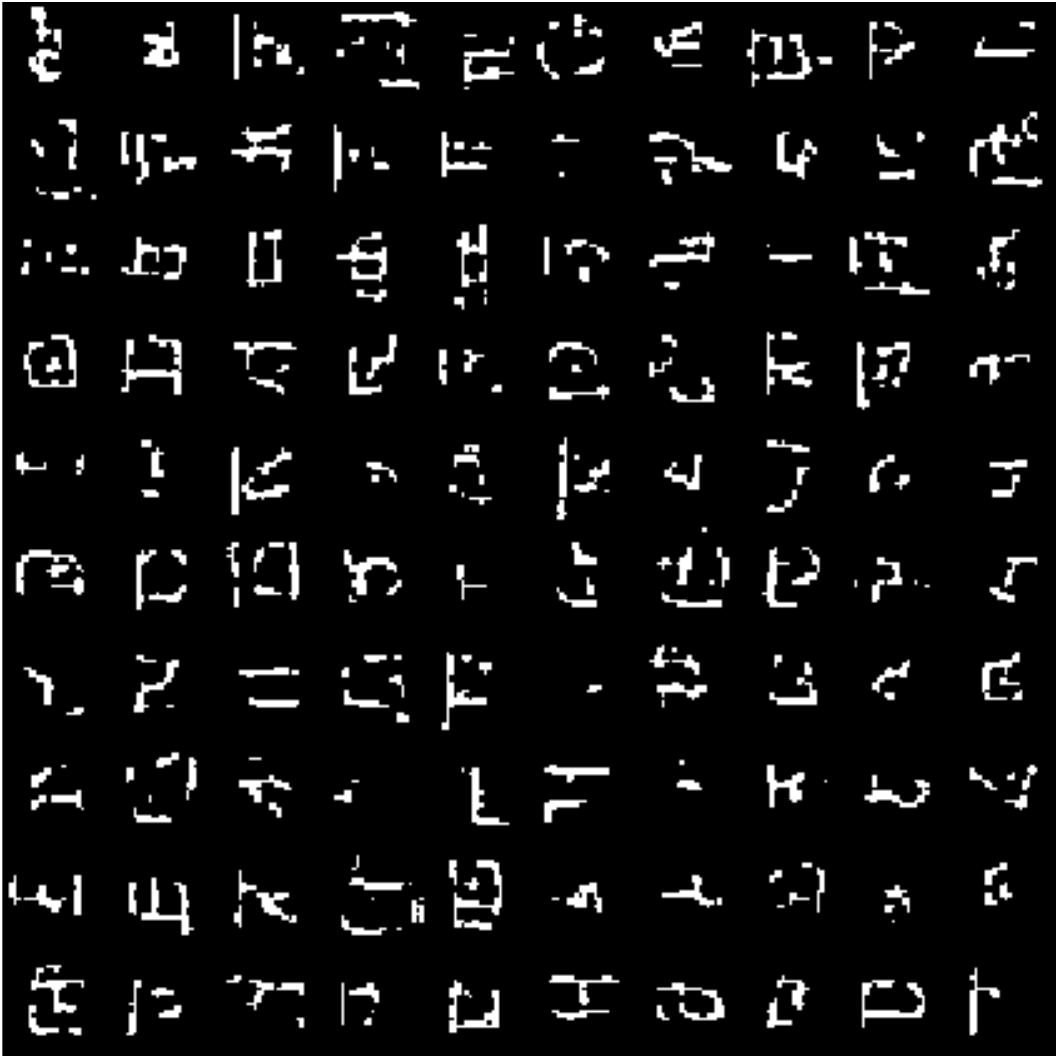}
    \caption{Random samples from IWAE-ConvFlow ($K=8$)}
  \end{subfigure}
  \begin{subfigure}[t]{0.245\textwidth}
    \includegraphics[width=\textwidth]{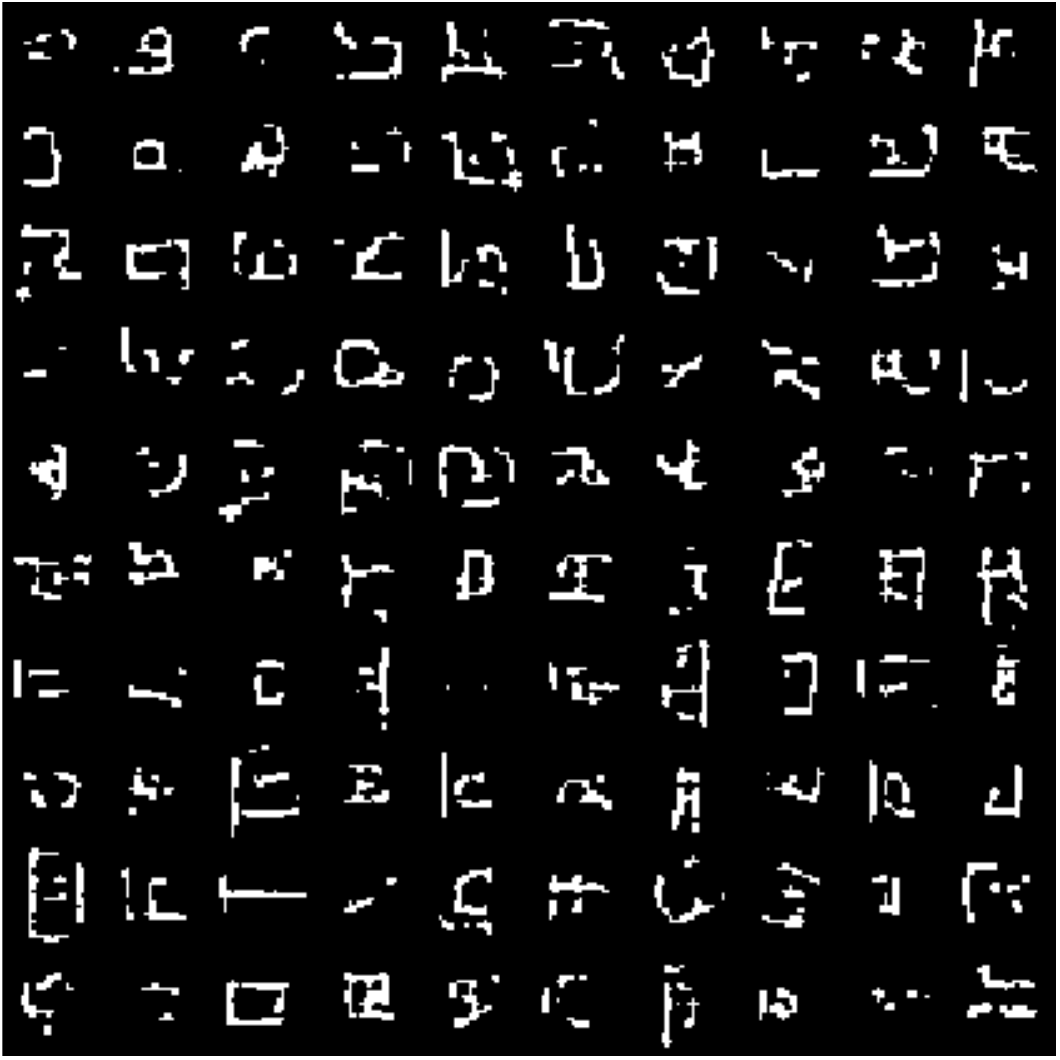}
    \caption{Random samples from IWAE-ConvFlow ($K=8$)}
  \end{subfigure}
  \begin{subfigure}[t]{0.245\textwidth}
    \includegraphics[width=\textwidth]{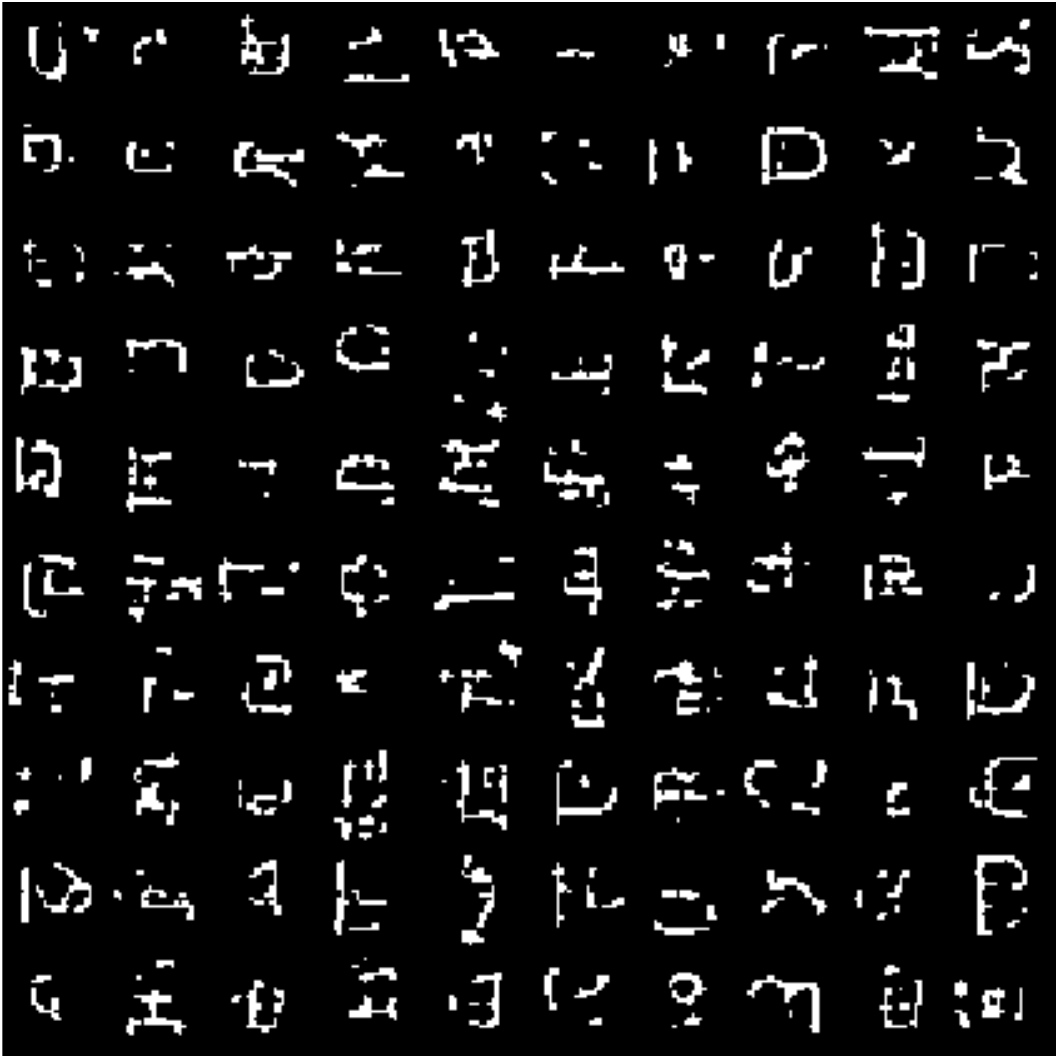}
    \caption{Random samples from IWAE-ConvFlow ($K=8$)}
  \end{subfigure}
  \caption{Training data and generated samples}
  \label{fig:samples}
\end{figure*}

After the models are trained, generative samples can be obtained by
feeding $z\sim N(0,I)$ to the learned generative model $G_1$ (or
$z_2\sim N(0,I)$ to $G_2$). Since higher log-likelihood estimates are
obtained on $G_2$, Figure \ref{fig:samples} shows three sets of random
generative samples from our proposed method trained with $G_2$ on both
MNIST and OMNIGLOT, compared to real samples from the training sets.
We observe the generated samples are visually consistent with the
training data.

\section{Conclusions}
\label{sec:conclusion}

This paper presents a simple and yet effective architecture to compose
normalizing flows based on 1d convolution on the input vectors.
ConvFlow takes advantage of the effective computation of convolution
to warp a simple density to the possibly complex target density, as
well as maintaining as few parameters as possible. To further
accommodate long range interactions among the dimensions, dilated
convolution is incorporated to the framework without increasing model
computational complexity. A Revert Layer is used to maximize the
opportunity that all dimensions get as much warping as
possible. Experimental results on inferring target complex density and
density estimation on generative modeling on real world handwritten
digits data demonstrates the strong performance of
ConvFlow. Particularly, density estimates on MNIST show significant
improvements over state-of-the-art methods, validating the power of
ConvFlow in warping multivariate densities. It remains an interesting
question to see how ConvFlows can be directly combined with powerful
observation models such as PixelRNN to further advance generative
modeling with tractable density evaluation. We hope to address these
challenges in future work.

\clearpage
\balance
\bibliographystyle{icml2018}
\bibliography{ref}

\appendix

\section{Conditions for Invertibility}
\label{app:condition}

The ConvFlow proposed in Section \ref{sec:convflow} is invertible, as
long as every term in the main diagonal of the Jacobian specified in
Eq. (\ref{eq:jacobian}) is non-zero, i.e., for all $i=1,2,...,d$,
\begin{align}
  w_1 u_i h'(\text{conv}(\vect z,w))+1\neq 0
\end{align}
where $u_i$ is the $i$-th entry of the scaling vector $\vect u$.  When
using $h(x)=\tanh(x)$, since $h'(x)=1-\tanh^2(x)\in [0,1]$, a
sufficient condition for invertibility is to ensure $w_1 u_i>
-1$. Thus a new scaling vector $\vect u'$ can be created from free
parameter $\vect u$ to satisfy the condition as
\begin{align}
  \vect u'=\begin{cases}
  \vect u &\text{if }w_1=0\\
  \vect -\frac{1}{w_1}+\text{softplus}(\vect u) &\text{if }w_1>0\\
  \vect -\frac{1}{w_1}-\text{softplus}(\vect u) &\text{if }w_1<0
  \end{cases}
\end{align}
where $\text{softplus}(x)=\log(1+\exp(x))$. The above sufficient
condition works readily for other non-linearity functions $h$ ,
including sigmoid, softplus, rectifier(ReLU), leaky rectifier (Leaky
ReLU) and exponential linear unit (ELU), as all their gradients are
bounded in $[0,1]$.
\end{document}